\documentclass{article}

\usepackage{PRIMEarxiv}

\usepackage[utf8]{inputenc} 
\usepackage[T1]{fontenc}    
\usepackage{hyperref}       
\usepackage{url}            
\usepackage{booktabs}       
\usepackage{amsfonts}       
\usepackage{nicefrac}       
\usepackage{microtype}      
\usepackage{lipsum}
\usepackage{amsmath}
\usepackage{siunitx}
\usepackage{amssymb}
\usepackage{bm}
\usepackage{algpseudocode}
\usepackage[ruled,vlined]{algorithm2e}
\usepackage[table]{xcolor}
\usepackage{amsthm}
\usepackage{fancyhdr}       
\usepackage{graphicx}       
\graphicspath{{media/}}     
\usepackage{schemata}
\usepackage{subcaption}
\newtheorem{remark}{Remark}
\usepackage{caption}

\renewcommand{\thefootnote}{\fnsymbol{footnote}}

\title{AB-PINNs: Adaptive-Basis Physics-Informed Neural Networks for Residual-Driven Domain Decomposition}

\author{
 Jonah Botvinick-Greenhouse$^{\dagger,*}$\\
  Center for Applied Mathematics \\
  Cornell University\\
  Ithaca, NY, 14850\\
  \texttt{jrb482@cornell.edu}\\  
  \AND   
 Wael H. Ali\\
  Mitsubishi Electric Research Laboratories\\
  Cambridge, MA, 02139\\
  \texttt{wali@merl.com}\\
  \AND
 Mouhacine Benosman\\
  Amazon Robotics\\
  North Reading, MA, 01864\\
  \texttt{m\_benosman@ieee.org}\\
  \AND
 Saviz Mowlavi$^*$\\
  Mitsubishi Electric Research Laboratories\\
  Cambridge, MA, 02139\\
  \texttt{mowlavi@merl.com}\\
}

\date{\today}

\begin{document}

\maketitle

\footnotetext[2]{This work was partially completed during an internship at Mitsubishi Electric Research Laboratories (MERL)}
\footnotetext[1]{Corresponding authors}
\renewcommand{\thefootnote}{\arabic{footnote}} 

\begin{abstract}
We introduce adaptive-basis physics-informed neural networks (AB-PINNs), a novel approach to domain decomposition for training PINNs in which existing subdomains dynamically adapt to the intrinsic features of the unknown solution. Drawing inspiration from classical mesh refinement techniques, we also modify the domain decomposition on-the-fly throughout training by introducing new subdomains in regions of high residual loss, thereby providing additional expressive power where the solution of the differential equation is challenging to represent. Our flexible approach to domain decomposition is well-suited for multiscale problems, as different subdomains can learn to capture different scales of the underlying solution. Moreover, the ability to introduce new subdomains during training helps prevent convergence to unwanted local minima and can reduce the need for extensive hyperparameter tuning compared to static domain decomposition approaches. Throughout, we present comprehensive numerical results which demonstrate the effectiveness of AB-PINNs at solving a variety of complex multiscale partial differential equations.
\end{abstract}


\section{Introduction}
Forward and inverse problems related to partial differential equations (PDEs) play a crucial role in many physical and biological applications, including  climate modeling, medical imaging, and heat transfer, to name a few. While classical techniques for solving PDEs, e.g., the finite difference and finite element methods \cite{li2017numerical}, have led to remarkable scientific advances, these approaches also require careful setup for inverse modeling via adjoint-based optimization \cite{plessix2006review}, are nontrivial to adapt to domains with complex geometry, rely on a fixed mesh, and struggle in high dimensions. 

In recent years, physics-informed neural networks (PINNs) \cite{raissi2019physics} have emerged as an alternative approach capable of overcoming some of these challenges. PINNs represent the unknown PDE solution via a deep neural network and are trained to minimize a residual loss which enforces the PDE constraint throughout the problem domain. Thus, PINNs are mesh-free, can be easily deployed over domains with complex geometry, and directly generalize to both parametric forward and inverse problems. Notably, one can simply incorporate a data-matching term in the loss function to solve inverse problems with PINNs, while classical adjoint-based optimization typically requires many evaluations of a costly forward model. Since their introduction, PINNs have successfully modeled various physical phenomena, including heat transfer \cite{cai2021physicsheat}, wave propagation \cite{moseley2020solving}, and fluid flows \cite{cai2021physics}.

While classical numerical schemes come with convergence guarantees, PINNs rely on neural network training of a nonconvex loss function. Though some theory for PINNs has been developed \cite{shin2020convergence,mishra2023estimates}, in general no performance guarantees are available, and it is well-known that PINNs often struggle to converge when the underlying problem exhibits high-frequency and multiscale behavior. Inspired by classical techniques for solving multiscale PDEs \cite{graham2007domain}, one promising direction for simplifying the PINN optimization involves performing domain decomposition. The core idea of domain decomposition is to partition the spatio-temporal domain into many smaller subdomains and to solve a relatively simple problem within each subdomain, while also ensuring consistent interface conditions between subdomains are satisfied. Several PINN-based domain decomposition methodologies  have already been developed \cite{li2019d3m,li2020deep,jagtap2020extended,liu2023cv,figueres2025pinn,shukla2021parallel,kopanivcakova2024enhancing}; see \cite{klawonn2024machine} for a comprehensive review.

Many of these methods involve training a separate neural network within each subdomain and enforcing interface conditions as a soft constraint, which requires careful construction of the loss function and can result in discontinuities of the learned PDE solution. The finite-basis physics-informed neural network (FBPINN) automatically enforces interface conditions by the construction of a suitable PINN solution ansatz \cite{moseley2023finite}. This method further simplifies the training loss and ensures continuity of the predicted PDE solution. In particular, the FBPINN ansatz is given by a summation of local solutions, each consisting of a product between a compactly supported window function and a subnetwork composed with a normalizing transformation mapping the support of the window function into reasonable inputs for effective subnetwork training. The support of each window function thereby defines a subdomain and each subnetwork is primarily responsible for learning the solution within its corresponding subdomain. In regions where subdomains overlap, multiple subnetworks contribute to the solution. FBPINNs have shown significant promise for learning the solution of high-frequency and multiscale PDEs where conventional PINN approaches struggle. Due to the individual subdomain normalization, each subnetwork is capable of representing a local solution using relatively few parameters. Moreover, the evaluation of all subnetworks can be vectorized, which makes the approach highly efficient at learning the unknown PDE solution.

In the FBPINN framework, the domain decomposition is determined before training begins, is fixed throughout training, and the only tunable parameters are the subnetworks' weights and biases. While adaptive mesh refinement has played a critical role in classical numerical methods for solving PDEs \cite{berger1984adaptive,burgarelli2006new,tang2021review}, there has not yet been extensive work on extending these ideas to PINN-based domain decomposition methodologies. For multiscale problems with local and sharp gradients, it is unlikely that a uniform domain decomposition is optimal for learning the underlying PDE solution. In particular, a decomposition which refines itself in areas of high PDE residual, i.e., where the solution is challenging to learn, may lead to faster convergence. 

While the implementation of adaptive mesh strategies for classical numerical methods can be complex and often requires numerous forward solves at different mesh resolutions, the simple nature of PINN training and flexibility of the FBPINN solution ansatz \eqref{eq:FBPINN} opens the door for dynamic PINN-based domain decomposition strategies which adapt throughout training. In this work, we propose adaptive-basis physics-informed neural networks (AB-PINNs), which have the following key features.
\begin{enumerate}
    \item \textbf{Local adaptivity of subdomains:} Each subdomain is parameterized as the support of a radial basis function (RBF) composed with a tunable affine transformation. Thus, during training the center and spread of each subdomain evolve to fit the geometry of the learned solution; see Section \ref{subsec:coords} for further discussion.
    \item \textbf{Residual-based addition of new subdomains:} During training, we introduce new adaptive subdomains  on-the-fly in regions of high PDE residual. This provides additional expressive power in regions where the solution is challenging to represent and prevents the PINN from getting stuck at a local minima; see Section~\ref{subsec:addition} for further discussion.
\item \textbf{Global network:} Different from the FBPINN framework, our model also makes use of a global network defined over the entire problem domain. The global network  helps learn the low frequency behavior of the PDE and ensures that the solution can still be learned in regions of the domain where subnetworks are not defined. 
\end{enumerate}

In the setting of large language models, the popular DeepSeekMoE architecture similarly makes use of global networks which capture common knowledge, while individual subnetworks are responsible for more specialized tasks \cite{dai2024deepseekmoe}. Moreover, while previous works have studied adaptive PINN-based domain decomposition strategies \cite{hu2023augmented,bischof2022mixture, stiller2020large}, our framework is the first to also consider the addition of new subdomains during the training procedure to target regions of high residual loss. In static decompositions such as FBPINNs, the optimal number and placement of subdomains are unknown without prior knowledge of the underlying solution, often requiring many trials with different subdomain initializations to obtain an accurate model.

The rest of the paper is structured as follows. Section \ref{sec:background} reviews background material including PINNs, FBPINNs, and related works. Section \ref{sec:AB-PINN} then introduces our proposed method, AB-PINNs, in full detail. Comprehensive numerical tests which demonstrate the effectiveness of our method are presented in Section \ref{sec:numerics}. Finally, conclusions follow in Section \ref{sec:conclusions}.

\section{Background}\label{sec:background}
In this section, we present background material which provides the context and motivation for our proposed method of adaptive domain decomposition in PINNs. In Section \ref{subsec:PINNs}, we introduce the standard PINN framework for solving partial differential equations. Sections \ref{subsec:weights}, \ref{subsec:decomp}, and \ref{subsec:architecture} then explore several innovations, e.g., adaptive sampling, domain decompositions, and modified architectures, which have since been used to improve the accuracy and convergence of the standard PINN. 

\subsection{Physics-Informed Neural Networks (PINNs)}\label{subsec:PINNs}
We now introduce the standard PINN framework for solving general partial differential equations of the form
\begin{align}
   \mathcal{D}[u](x) &= f(x), \qquad x\in \Omega, \label{eq:inside}\\
   \mathcal{B}[u](x) &= g(x), \qquad x\in \Gamma \subseteq \partial \Omega, \label{eq:outside}
\end{align}
where $\Omega\subseteq \mathbb{R}^d$ is a bounded domain, $\partial \Omega$ is the boundary, $\mathcal{D}$ is a differential operator, $\mathcal{B}$ is a boundary operator, $f$ is a forcing function, and $g$ is a boundary function. A PINN represents a solution $u:\Omega \to \mathbb{R}$ of the system \eqref{eq:inside}-\eqref{eq:outside} by a neural network $u_{\theta}:\Omega \to \mathbb{R}$, with tunable parameters $\theta\in \Theta \subseteq \mathbb{R}^p$. The network is then trained via gradient-based methods to reduce a suitable loss that encodes the interior PDE constraint \eqref{eq:inside}, as well as the boundary condition \eqref{eq:outside}. While the original PINN formulation \cite{raissi2019physics} enforces the boundary conditions as a soft constraint incorporated in the loss function, in many situations the boundary condition \eqref{eq:outside} can be directly enforced as a hard constraint \cite{dong2021method,wang2023expert}, which prevents the need to balance multiple terms in the loss and helps accelerate convergence. By leveraging knowledge of the boundary operator $\mathcal{B}$, in certain situations one may construct a suitable constraining operator $\Psi$ and instead regard $\hat{u}_{\theta}= \Psi[u_{\theta}]$ as the ansatz for solving the PDE, which automatically satisfies $\mathcal{B}[\hat{u}_{\theta}](x) = g(x) $, for all $x\in \Gamma \subseteq \partial \Omega$; see Section \ref{sec:numerics} for several examples of constraining operators. The resulting optimization problem and physics-informed loss function are then given by
\begin{equation}\label{eq:optimization}
    \min_{\theta\in \Theta}\mathcal{L}(\theta),\qquad \mathcal{L}(\theta) = \frac{1}{\text{vol}(\Omega)}\int_{\Omega} \mathcal{R}_{\theta}(x)\,\textrm{d}x,\qquad \mathcal{R}_{\theta}(x):= |\mathcal{D}[\hat{u}_{\theta}(x)] - f(x)|^2.
\end{equation}
In \eqref{eq:optimization}, $|\cdot|$ denotes the Euclidean 2-norm, and $\mathcal{R}_{\theta}:\Omega \to \mathbb{R}$ is the so-called \textit{PDE residual}. The residual $\mathcal{R}_{\theta}(x)$ is computed via automatic differentiation and determines how well the solution ansatz $\hat{u}_{\theta}$ satisfies \eqref{eq:inside} at a given point $x\in\Omega$. When the loss $\mathcal{L}(\theta)$ defined by \eqref{eq:optimization} is minimized to zero it holds that $\mathcal{R}_{\theta}(x) = 0$ for all $x\in \Omega$, thus ensuring that $\hat{u}_{\theta}$ solves \eqref{eq:inside}-\eqref{eq:outside}. In practice, the integral in \eqref{eq:optimization} cannot be analytically computed and one relies on the Monte--Carlo approximation 
\begin{equation}\label{eq:MC}
    \hat{\mathcal{L}}(\theta) = \frac{1}{N} \sum_{k=1}^N \mathcal{R}_{\theta}(x_k),
\end{equation} where $\{x_k\}_{k=1}^N$ are \textit{collocation points} sampled i.i.d. from $\text{Unif}(\Omega)$,  the uniform distribution over $\Omega$.  Using fixed collocation points throughout training may lead to overfitting the PDE residuals, and thus it is best practice to randomly resample the collocation points throughout training \cite{wang2023expert}. 

While the PINN framework described above is conceptually straightforward and simple to implement, in practice the approach can struggle to converge when the underlying differential equation exhibits complex, multiscale behavior. In recent years, researchers have identified several key modifications to the PINN training pipeline which significantly improve the solution accuracy; see \cite{wang2023expert} for a review of these developments. 
\subsection{Adaptive Sampling and Reweighted Residuals}\label{subsec:weights}

To improve the convergence of PINNs, several works utilize reweighted formulations of the loss \eqref{eq:optimization}. The key idea behind these strategies is to focus training on regions of the spatio-temporal domain where the PINN may be struggling to represent the underlying solution. More specifically, given a strictly positive density $\pi:\Omega \to \mathbb{R}$  one can instead consider the loss
\begin{equation}\label{eq:weighted1}
 \mathcal{L}_{\pi}(\theta)= \int_{\Omega}\mathcal{R}_{\theta}(x) \pi(x) \, \textrm{d}x.
\end{equation}
Clearly, if $\mathcal{L}_{\pi}(\theta) = 0$, then $\hat{u}_{\theta}$ still solves \eqref{eq:inside}-\eqref{eq:outside}. Writing down the Monte--Carlo approximation of \eqref{eq:weighted1} leads to a new finite-sample loss:
\begin{align}
  \hat{ \mathcal{L}}_{\pi}(\theta) = \frac{1}{N}\sum_{k=1}^N \mathcal{R}_{\theta}(x_k), \qquad x_k\sim \pi.\label{eq:MC2}
\end{align}
The strategy \eqref{eq:MC2}, which involves drawing the collocation points from the distribution $\pi$, has motivated numerous works to study optimal strategies for selecting $\pi$ in practice  \cite{wu2023comprehensive,gao2023failure,mao2023physics,jiao2024gaussian}. Most of these focus on adaptive strategies in which $\pi$ is determined using the current solution ansatz $\hat{u}_{\theta}$ and  PDE residual $\mathcal{R}_{\theta}$. A simple approach studied in \cite{wu2023comprehensive} involves selecting $\pi(x) \propto \mathcal{R}_{\theta}(x)+c,$ the motivation being that more collocation points should be sampled in regions of high PDE residual, thereby focusing the PINN to train where the underlying solution is challenging to represent. 

Another class of approaches avoids the need to directly sample from the distribution $\pi$ and instead assigns a weight $w_k > 0$ to each collocation point, i.e., 
\begin{equation}\label{eq:MC3}
    \hat{\mathcal{L}}_w(\theta) = \frac{1}{N}\sum_{k=1}^N w_k\mathcal{R}_{\theta}(x_k), \qquad x_k\sim \text{Unif}(\Omega).
\end{equation}
Note that if $w_k = \pi(x_k)\cdot \text{vol}(\Omega)$, then \eqref{eq:MC3}  also approximates \eqref{eq:weighted1} via importance sampling. Multiple approaches have been developed to assign pointwise weights $w_k$, either through direct gradient ascent or by defining or learning a residual-to-weight mapping, in order to guide PINN training toward regions exhibiting large PDE residuals \cite{mcclenny2023self,zhang2023dasa,anagnostopoulos2024residual,song2024loss}.
A different line of work~\cite{wang2024respecting} focuses on choosing adaptive residual weights which enforce temporal causality, such that the solution is learned sequentially over time starting from the initial condition. 

We remark that our proposed method for adaptive domain decomposition primarily involves changing the base neural network architecture and is compatible with any adaptive sampling or residual reweighting scheme, which are expected to improve performance.

\subsection{Finite-Basis Physics Informed Neural Networks} \label{subsec:decomp}

We now describe the finite-basis physics informed neural network (FBPINN) \cite{moseley2023finite} in detail. FBPINNs are a ``soft'' domain decomposition strategy for training PINNs, which does not involve learning interface conditions between subdomains during optimization. 
In particular, the FBPINN solution ansatz is given by 
\begin{equation}\label{eq:FBPINN}
  u_{\theta}(x) = \sum_{i=1}^K \phi_i(x) \texttt{NN}_i(r_i(x);\theta_i),  \qquad x\in \Omega,
\end{equation}
where $\Omega \subseteq \mathbb{R}^d$ is the problem domain,  $K$ is the number of subdomains,  $\phi_i$ is a smooth (effectively) compactly supported \textit{window function} defining the support of each subdomain, the \textit{subnetwork} $\texttt{NN}_i$ is a universal function approximator, $\theta_i\in \Theta\subseteq \mathbb{R}^p$ comprise its tunable parameters, and $r_i$ is a coordinate transformation normalizing the support of $\phi_i$ to a standard input range for effective neural network training. In \eqref{eq:FBPINN}, the window functions $\phi_i$ should be chosen to have overlapping supports and cover the full domain $\Omega$. The subnetwork $\texttt{NN}_i$ is primarily responsible for learning the solution within the support of $\phi_i$, and in any region where subdomains overlap the solution is given by the summation of multiple subnetwork contributions. For a visualization of the overlapping FBPINN subdomains, see \cite[Figure 2]{moseley2023finite}.

In Section \ref{sec:AB-PINN}, we make the parameters of the window functions $\phi_i$ and coordinate transformations $r_i$ tunable, enabling each subdomain to adapt to the geometry of the underlying PDE during training. Second, we modify the decomposition \eqref{eq:FBPINN} on-the-fly by introducing new subdomains during training in regions of high PDE residual. We show in Section \ref{sec:numerics} that both of these modifications significantly improve the convergence of the FBPINN framework and can substantially reduce the number of parameters needed to train an accurate FBPINN model for multiscale problems. 

\subsection{Modified Network Architecture}\label{subsec:architecture}
Several works have investigated how modifying the base neural network architecture can improve the performance of PINNs. While it is standard in many cases to use a standard multi-layer perceptron (MLP) with Glorot initialization \cite{glorot2010understanding}, modified MLPs \cite{wang2021understanding}, Kolmogorov Arnold Networks (KANs) \cite{liu2025kan}, and SIREN \cite{sitzmann2020implicit}, are variations which have exhibited improved performance in certain situations. Similar to the FBPINN setup, our approach can be viewed as a new neural network architecture used to train PINNs. While in this work, we parameterize each subnetwork $\texttt{NN}_i$ as an MLP, any of the previously mentioned architecures are also compatible with our framework.

\subsection{Related Work}

Three notable works \cite{hu2023augmented,bischof2022mixture, stiller2020large} study adaptive domain decompositions for training PINNs from a mixture-of-experts perspective \cite{shazeer2017outrageously}, using a gating network to parameterize the subdomains. Importantly, our approach uses tunable radial basis functions to define the domain decomposition, which easily allows the user to (1) introduce new subdomains in regions of high PDE residual, and (2) define simple coordinate transformations for each subdomain which are crucial for efficient subnetwork training. We remark that neither of these key capabilities is straightforward when the domain decomposition is parameterized by a gating network. Moreover, while various works have studied the use of RBF networks with adaptive basis function parameters for solving PDEs \cite{shao2025solving, ramabathiran2021spinn,wang2023solving, dwivedi2025kernel}, these approaches do not consider the use of subnetworks with coordinate normalization to improve expressivity of the model on the support of each basis function. A different direction uses a PINN-based residual to perform adaptive mesh refinement for a classical numerical PDE solver \cite{zhu2024unstructured}. Finally, we note that several works have also studied neural networks with adaptive architectures to enhance PINN performance outside of the context of domain decomposition \cite{jagtap2020locally,wang2023learning,wang2024piratenets,rigas2024adaptive}.  

\section{Adaptive-Basis Physics-Informed Neural Networks (AB-PINNs)}\label{sec:AB-PINN}
In this Section we describe our proposed AB-PINN model in full detail. Section \ref{subsec:coords} first discusses how we parameterize each trainable subdomain. In Section \ref{subsec:addition}, we discuss our method for introducing new subdomains in regions of high PDE residual. Section \ref{subsec:periodicity} then discusses how AB-PINNs are made compatible with Fourier embeddings to enforce periodicity of the predicted PDE solution, which is required specifically for the PDEs considered in Sections \ref{subsec:adv}, \ref{subsec:ac}, and \ref{subsec:kdv}. Finally, in Section \ref{subsec:implementation} we discuss some practical implementation details for the AB-PINN model.

\subsection{Learnable Subdomains}\label{subsec:coords}
We now describe the base architecture of AB-PINNs. First, we modify \eqref{eq:FBPINN} such that the window function $\phi_i$ is comprised of tunable parameters, which in turn define the coordinate transformation $r_i.$ While there are many ways to parameterize both the window functions $\phi_i$ and resulting coordinate transformations $r_i$, we choose a simple representation by constructing the functions with a tunable affine transformation. In particular, we define some reference window function $\psi:\Omega \to \mathbb{R}$ and define each tunable window function and coordinate transformation by 
\begin{equation}\label{eq:coords}
    \phi_i(x):= (\psi \circ r_i)(x),\qquad r_i(x):= L_i^{\top}(x-\mu_i), \qquad L_i \in \mathbb{R}^{d\times d}, \qquad \mu_i \in \mathbb{R}^d.
\end{equation}
We choose $L_i$ as a lower-triangular matrix with strictly positive diagonal, where the parameters of each  $L_i\in \mathbb{R}^{d\times d}$ and $\mu_i\in \mathbb{R}^d$ are learnable. For notational simplicity, we write $\beta_i$ to denote this vector of tunable parameters corresponding to $r_i$. Notably, when $\psi$ is chosen as a \textit{radial basis function} (RBF), i.e., its output depends only on the norm of the input vector through a function $\hat{\psi}:[0,\infty) \to \mathbb{R}$, we have that
\begin{equation}\label{eq:RBF}
    \phi_i(x) = \hat{\psi}(|r_i(x)|) = \hat{\psi}( \|x-\mu_i\|_{M_i}), \qquad M_i = L_iL_i^{\top}\in \mathbb{R}^{d\times d},\qquad \|x\|_M:= \sqrt{x^{\top}M x}.
\end{equation}
The window functions $\phi_i$ thus become RBFs centered at $\mu_i$ and stretched along the eigenvectors of $M_i$. Importantly, our parameterization \eqref{eq:coords} can represent any shift $\mu_i$, as well as stretches induced by any norm of the form $\|\cdot \|_{M_i}$. Indeed, the Cholesky decomposition guarantees that any symmetric positive-definite matrix $M_i$ admits a unique decomposition of the form $L_iL_i^{\top}$ where $L_i$ is a lower-triangular matrix with strictly positive diagonal entries. 

\begin{remark}\label{ex:1}
    Consider the case where $\psi(x) = \exp(-|x|^2/2).$ Then, by \eqref{eq:RBF} we have 
    \begin{align*}
        \phi_i(x) = \exp\bigg(-\frac{1}{2}(x-\mu_i)^{\top} L_i L_i^{\top}(x-\mu_i)  \bigg) = \exp\bigg(-\frac{1}{2}(x-\mu_i)^{\top} \Sigma_i^{-1} (x-\mu_i)  \bigg), \quad \Sigma_i:= \big(L_i L_i^{\top}\big)^{-1}\in \mathbb{R}^{d\times d},
    \end{align*}
    which is proportional to a multivariate Gaussian distribution with mean $\mu_i$ and covariance $\Sigma_i$, both of which are fully tunable during training. In this setup, the coordinate transformation $r_i:\Omega \to \Omega$ is a so-called whitening transformation that maps samples from $\mathcal{N}(\mu_i,\Sigma_i)$ to samples from $\mathcal{N}(0, I).$
\end{remark}
 In practice, while one can choose any radial basis function $\psi$, it is helpful to select a master window function $\psi$ such that the support of $\psi$ contains inputs in the range that is best-suited for neural network training, e.g., samples with mean zero and unit variance. By selecting $\psi$ in this way, the coordinate transformation $r_i$ maps the support of the window function $\phi_i$ into this range, thus providing $\texttt{NN}_i(r_i(x))$ the best capability to represent the local solution on the support of $\phi_i.$ For example, for a one-dimensional domain, choosing $\psi$ to be compactly supported on $[-1,1]$ automatically ensures that $r_i$ normalizes the support of the window function $\phi_i$ to $[-1,1]$, which corresponds to the strategy used in \cite{moseley2023finite}. In Figure \ref{fig:basis_fig}, we compare the performance of several AB-PINN and FBPINN models with different choices of $\psi.$  
 
 Given that the window functions $\phi_i$ are now fully adaptive, one cannot guarantee that their support will entirely cover the domain $\Omega$ during training. To prevent situations in which parts of the domain are left outside the effective support of the window functions, and to promote the propagation of information between sub-domains, we also include a fixed global network as part of our parameterization. This network helps learn the low-frequency behavior of the unknown PDE solution and can also reduce the number of subdomains necessary for successful training.

Everything together, we can write our base architecture for the AB-PINN as 
 \begin{equation}\label{eq:AB-PINN}
     u(x;\theta,\beta) = \texttt{NN}_0(x;\theta_0)+\sum_{i=1}^K\phi_i(x;\beta_i)\texttt{NN}_i(r_i(x;\beta_i);\theta_i). 
 \end{equation}
 Note that the only differences between \eqref{eq:AB-PINN} and \eqref{eq:FBPINN} are the parameterization of the window functions $\phi_i$ and coordinate transformations $r_i$, as well as the addition of the global network $\texttt{NN}_0$.

\subsection{Residual-Based Subdomain Addition}\label{subsec:addition}

While \eqref{eq:AB-PINN} constitutes the base AB-PINN architecture, we also modify \eqref{eq:AB-PINN} on-the-fly during training to introduce new subdomains in regions of high PDE residual. This involves constructing a new window function $\phi_{K+1}$ with parameters $\beta_{K+1}$ and a new subnetwork $\texttt{NN}_{K+1}$ with parameters $\theta_{K+1}$. While the subnetwork parameters $\theta_{K+1}$ are initialized according to a standard Glorot scheme \cite{glorot2010understanding}, the choice of $\beta_{K+1}$, i.e., the center and spread of the new subdomain, plays a crucial role in the PINN training. First, the location of the new subdomain is adaptively determined as $\mu_{K+1}\sim \pi$, where $\pi$ is the distribution proportional to the PDE residual $\mathcal{R}_{\theta}$; see \eqref{eq:optimization}. Thus, the center of the new subdomain is likely to be placed in a region of high PDE residual, where the current solution ansatz is struggling to represent the true solution. Note the similarity between this strategy and the adaptive sampling strategies discussed in Section~\ref{subsec:weights}. The spread of the new subdomain, determined by the lower-triangular matrix $L_{K+1}$, is simply a user-specified hyper-parameter which indicates the fraction of the domain the new subnetwork is responsible for. After the new subdomain is placed, both its center and spread continue to dynamically evolve during training, as described in Section \ref{subsec:coords}. For the remainder of the paper, we will write AB-PINN+ as shorthand for an AB-PINN model which uses residual-based subdomain addition.

In our numerical experiments, we observe that the addition of new subdomains in regions of high PDE residual can be useful for avoiding local minima and accelerating convergence; see Section \ref{subsec:ac}. Moreover, in certain situations, the domain decomposition learned via residual-based subdomain addition can provide significant benefits over a decomposition arrived at by initializing all subdomains at the start of training; see Section \ref{subsec:poisson}. There are also notable computational savings gained by introducing the subdomains throughout training, rather than starting with all subdomains at the beginning of training.

\subsection{Enforcing Periodicity in AB-PINNs}\label{subsec:periodicity}

As with any PINN architecture, a suitable constraining operator can be applied to enforce a Dirichlet or Neumann boundary or initial condition as a hard constraint in AB-PINNs; see Section \ref{subsec:PINNs}. In several of our numerical experiments the underlying PDE is also subject to periodic spatial boundary conditions. While these boundary conditions can be imposed as a soft constraint in the loss function \cite{raissi2019physics}, it is also possible to use an input-coordinate Fourier embedding to enforce periodicity as a hard constraint \cite{wang2023expert}. We illustrate how this strategy is made compatible with AB-PINNs in a simple one-dimensional case that generalizes to higher dimensions.

For a one-dimensional spatial input $x\in [-1,1]$, the Fourier embedding is given by the map $x\mapsto (\cos(\pi x),\sin(\pi x))\in \mathbb{S}^1\subseteq \mathbb{R}^2$, where $\mathbb{S}^1$ is the unit circle. By defining a neural network that acts on the embedded space $u_{\theta}:\mathbb{R}^2\to \mathbb{R}$, one can ensure that the function $v_{\theta}:[-1,1]\to \mathbb{R}$ defined by $v_{\theta}(x) = u_{\theta}(\cos(\pi x),\sin(\pi x))$ satisfies the periodic condition $v_{\theta}(-1) = v_{\theta}(1).$ In our numerical experiments, we model $u_{\theta}$ as an AB-PINN defined on $\mathbb{R}^2$ with the centers $\mu_i \in \mathbb{R}^2$ constrained to remain inside $\mathbb{S}^1$ via the Fourier embedding, while the lower-triangular matrices $L_i\in \mathbb{R}^{2\times 2}$ are parameterized as before.  Thus, the AB-PINN subdomain decomposition occurs within the embedded coordinates rather than in the original spatial domain, automatically ensuring periodicity of the window functions $\phi_i$ in the original domain. 

\subsection{Additional Implementation Details}\label{subsec:implementation}
 During training of the AB-PINN architecture \eqref{eq:AB-PINN}, all window functions $\{\phi_i\}_{i=1}^K$, coordinate transformations $\{r_i\}_{i=1}^K$, and subnetworks $\{\texttt{NN}_i\}_{i=1}^K$ are simultaneously evaluated in a single vectorized pass, which makes the Ansatz \eqref{eq:AB-PINN} practical for complex problems consisting of many subdomains. One of the primary reasons FBPINNs are so efficient is that each subnetwork only needs to be evaluated on collocation points falling within its respective subdomain. While for simplicity we evaluate each AB-PINN subnetwork on all collocation points, existing work from the mixture-of-experts community can provide methods for selective collocation point evaluation which is expected to improve efficiency~\cite{mu2025comprehensive,gale2023megablocks}.

 Motivated by the observation that sometimes small perturbations of a window function can lead to significant changes in the predicted PDE solution, we also find it helpful to manually freeze the learning rates of the tunable window function parameters $\mu_i$ and $L_i$ after a fixed number of iterations have elapsed. This effectively separates the AB-PINN training into two distinct phases: 
 \begin{enumerate}
     \item \textbf{Learning the solution features:} The window functions adapt, alongside all subnetwork and global network parameters, to learn the distribution of the underlying solution features over the domain.
     \item \textbf{Fine-tuning:} The window function parameters are frozen, and the subnetworks and global networks continue training until convergence.
 \end{enumerate}
We have empirically observed that the addition of the fine-tuning stage  improves the AB-PINN training stability and prediction accuracy.

\section{Numerical Experiments}\label{sec:numerics}
\subsection{Overview of Numerical Experiments}
\begin{figure}[h!]
    \centering

   \subfloat[Chirp waveform]{\includegraphics[width=.3\textwidth]{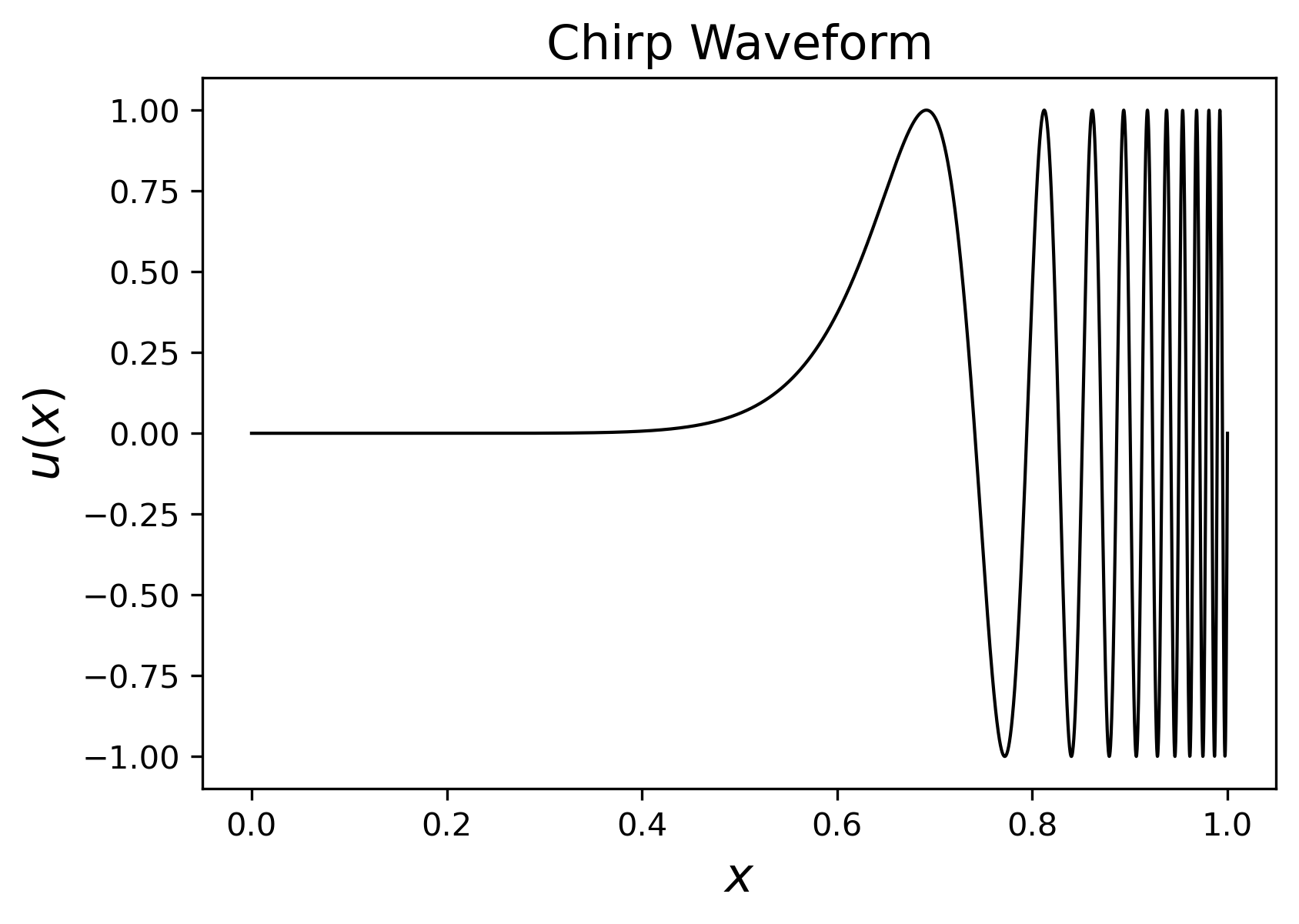}\label{fig:chirpref}}  \quad
  \subfloat[Advection equation ]{\includegraphics[width=.3\textwidth]{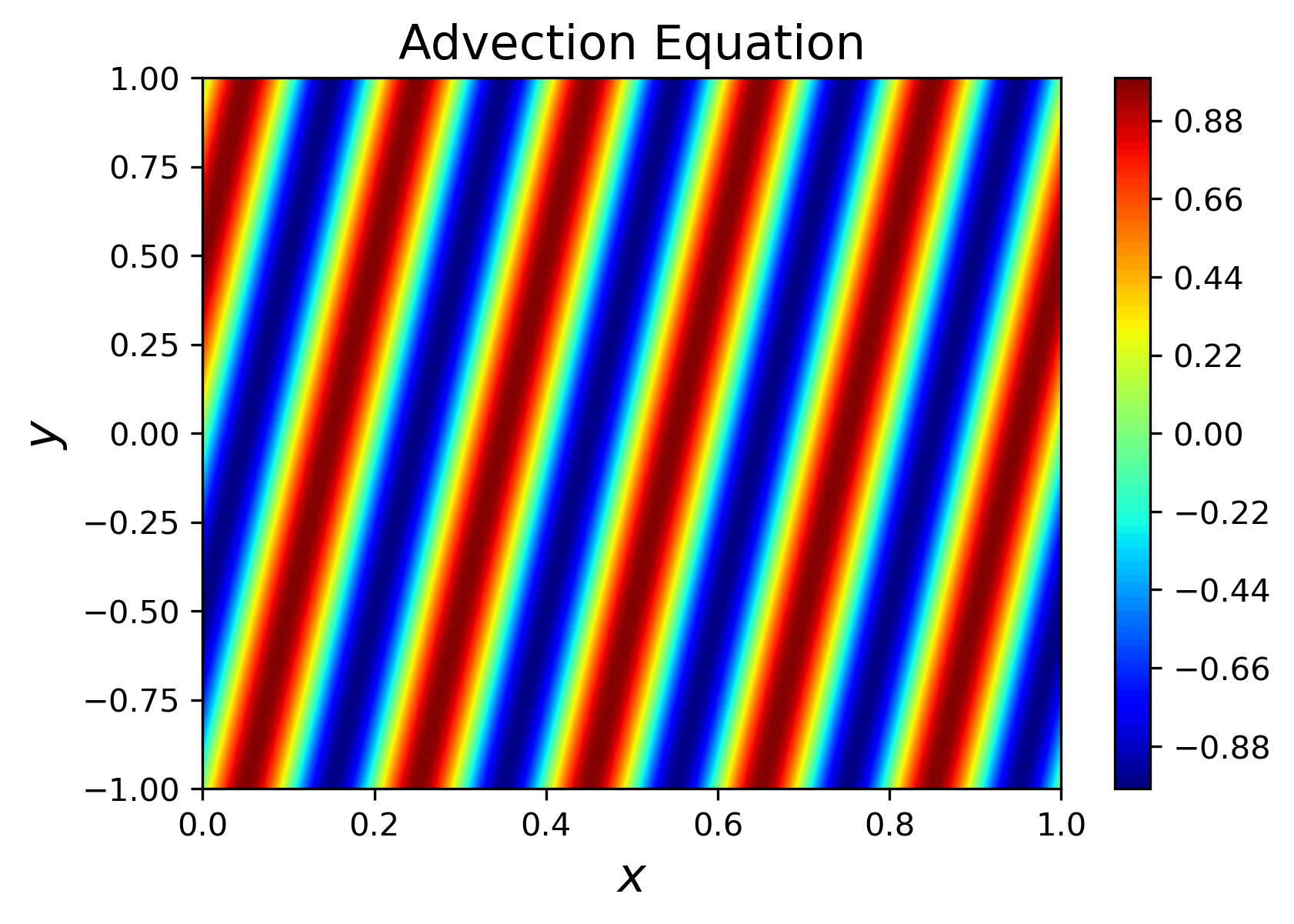}\label{fig:advref}}\quad
  \subfloat[Helmholtz equation ] {\includegraphics[width=.3\textwidth]{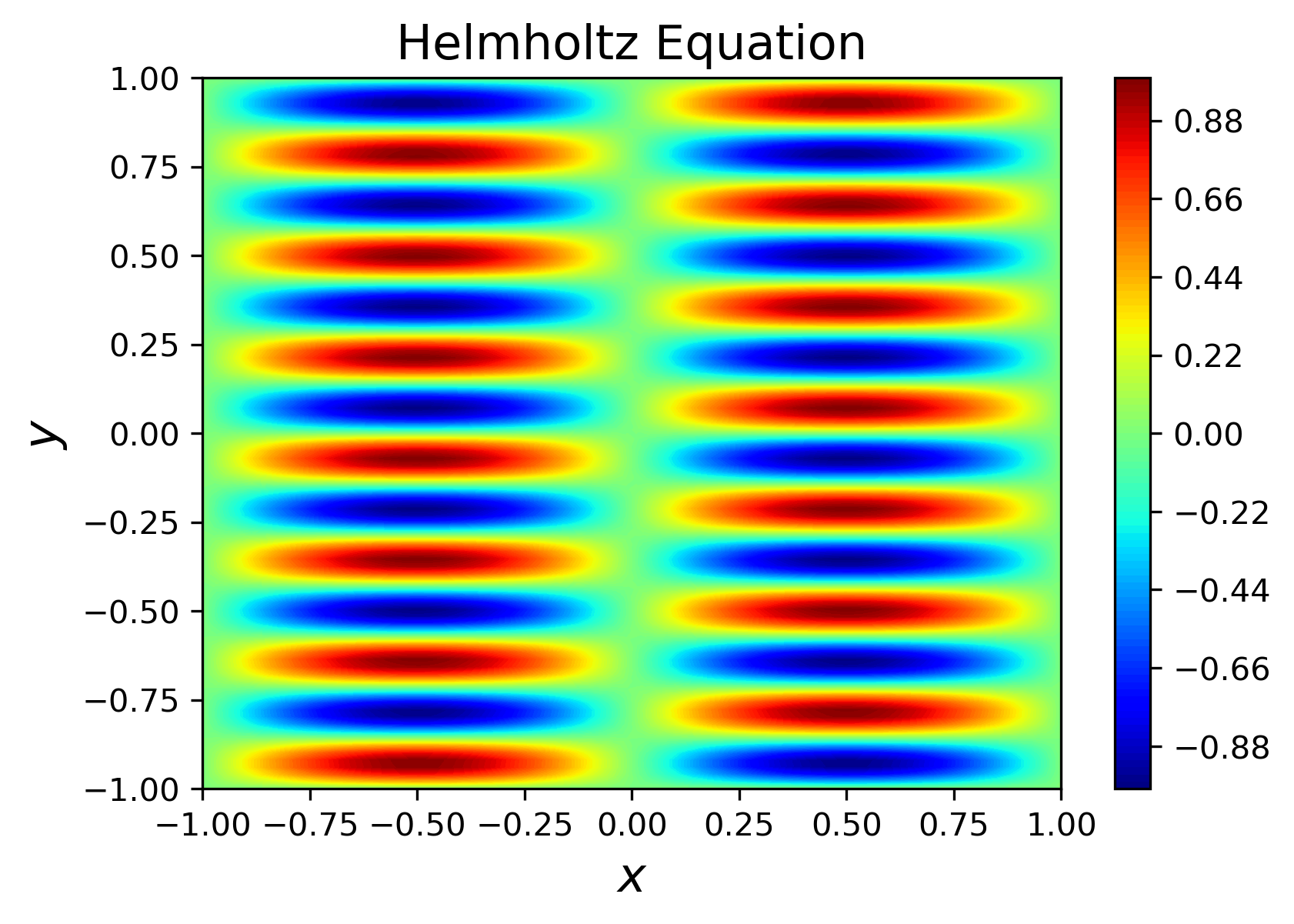}\label{fig:helmref}} 

 \subfloat[Forced Poisson equation]{\includegraphics[width=.3\textwidth]{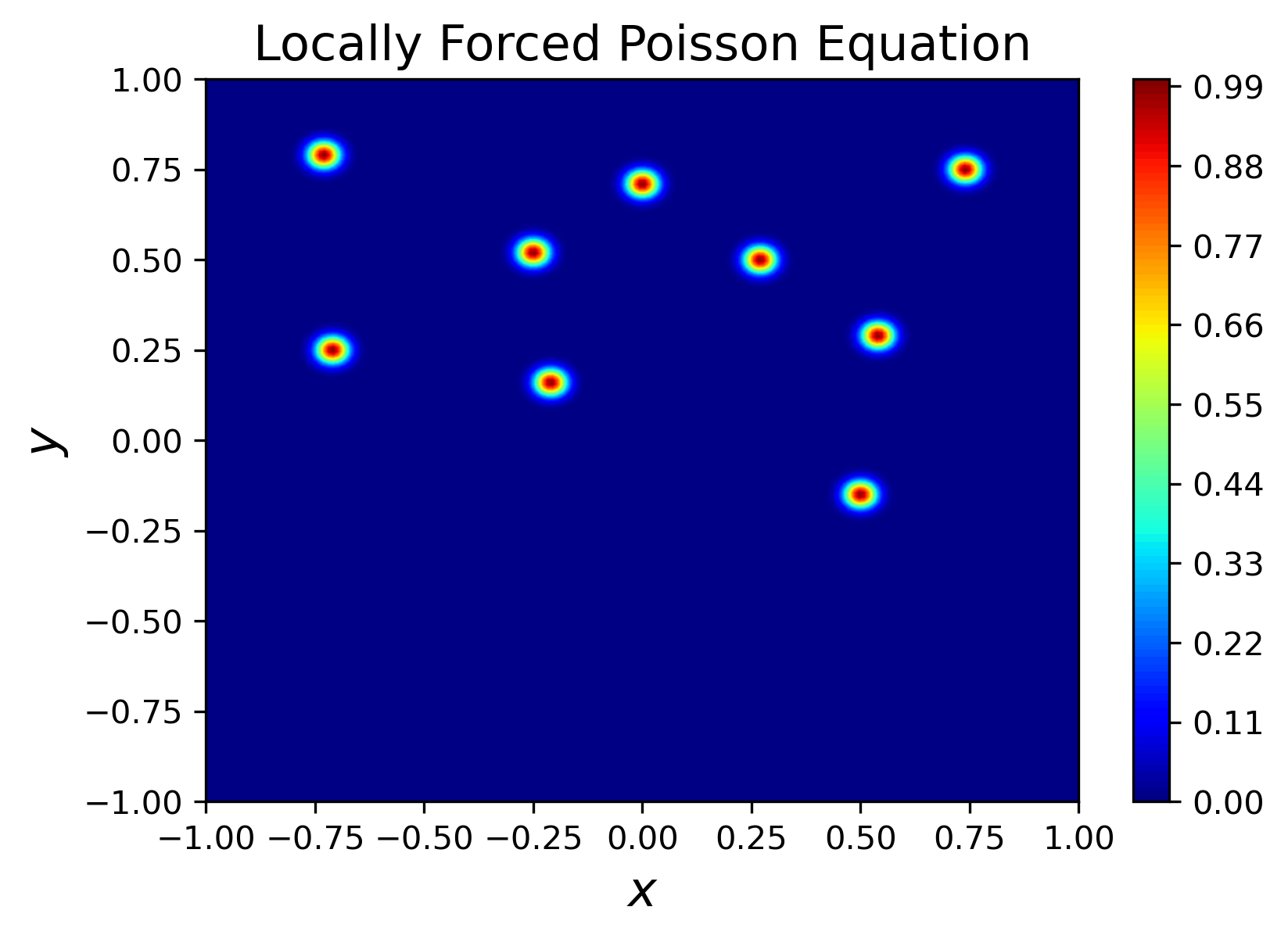}\label{fig:poissref}} \quad
   \subfloat[Allen--Cahn equation ]{\includegraphics[width=.3\textwidth]{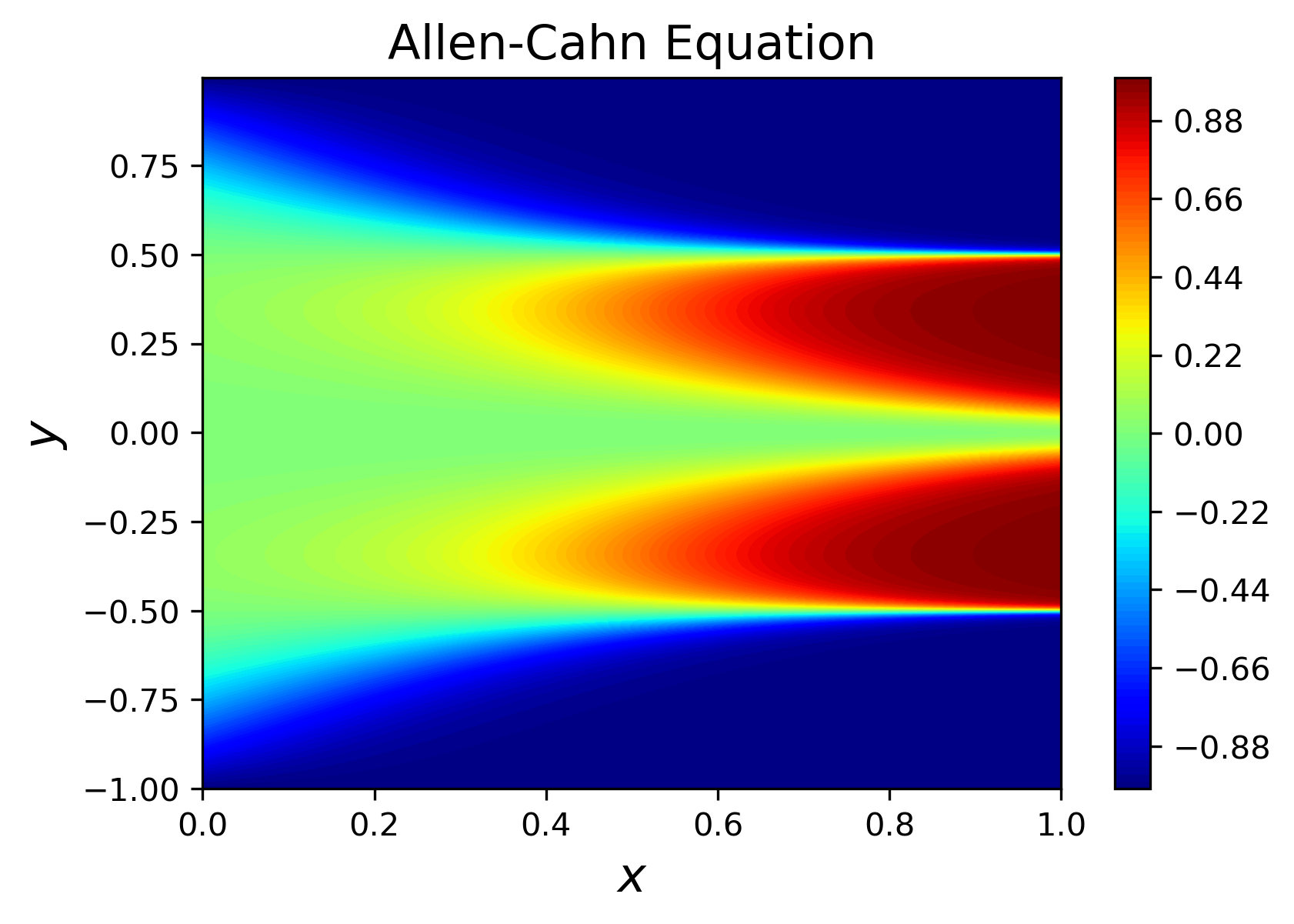}\label{fig:cahnref}} \quad 
    \subfloat[Korteweg--De Vries equation]{\includegraphics[width=.3\textwidth]{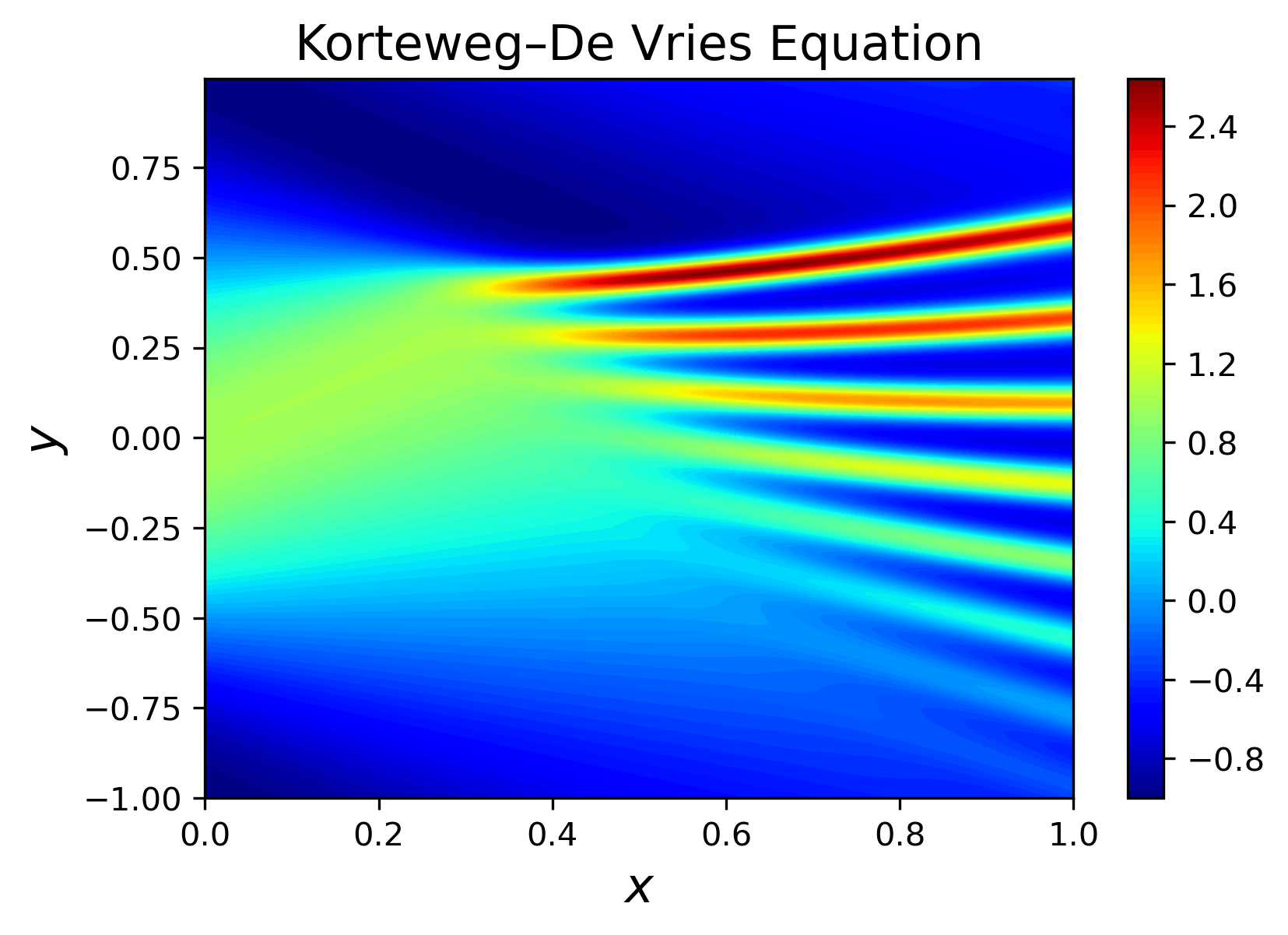}\label{fig:kdvref}}
  \label{fig:example_refs}
   \caption{Reference solutions for the test cases studied in this paper.} \label{fig:refsols}
\end{figure}

In this section, we present comprehensive numerical results which showcase the effectiveness of AB-PINNs at solving complex multiscale differential equations compared to both standard PINNs and FBPINNs. 
Throughout, all PINN models are trained on the CPU of a MacBook Air using the Apple M3 chip.  Figure \ref{fig:refsols} displays the ground truth reference solutions for the test cases we consider, including a multiscale chirp waveform, the advection equation, a Helmholtz equation, a locally forced Poisson equation, the Allen--Cahn equation, and the Korteweg--De Vries (KdV) equation. The reference solutions for the Allen--Cahn and KdV equations are approximated using a high-accuracy spectral method, while the solutions for the remaining differential equations are analytically available.

We begin our numerical experiments in Section \ref{subsec:local_adaptive_exp} by considering a fixed AB-PINN architecture without the subdomain addition described in Section \ref{subsec:addition}. The only difference between this architecture and the FBPINN is the adaptivity of the window functions and the inclusion of a global network for learning low-frequency components of the solution. Under this setup, we show in Table \ref{tab:loc} that AB-PINNs can improve upon the accuracy of FBPINNs by several orders of magnitude when solving multiscale problems whose solution features are not uniformly distributed across the spatio-temporal domain. The only example we consider where the FBPINN has higher accuracy than the AB-PINN is in approximating a sine wave solution. In this case, the uniform configuration of the FBPINN window functions is likely already optimal and any perturbation from this configuration learned by the AB-PINN may cause a disadvantage. 

We then go on to show in Section \ref{subsec:global_adaptive_exp} how the residual-based subdomain addition can be used to avoid local minima during training and deliver needed expressive power in regions of the spatio-temporal domain where the PINN is struggling to converge. In these experiments, we begin with a standard PINN architecture and add AB-PINN+ subdomains throughout training in regions of high PDE residual. The results for these experiments are summarized in Figure \ref{tab:glob}, revealing that adding subdomains during training steers the AB-PINN+ away from bad local minima and toward the correct solution. We also study how the choice of window function affects the performance of AB-PINNs. Moreover, we showcase the differences in the learned domain decomposition either when using the residual-based subdomain addition (AB-PINN+) or when relying solely on the local adaptivity of existing subdomains (AB-PINN).

\begin{table}[h!]
\centering
\begin{subtable}[t]{\textwidth}
\centering
\begin{tabular}{|c|c|c|c|c|}
\hline
 & Chirp Waveform &Sine Wave & Helmholtz equation & Advection equation \\ \hline Standard PINN& $6.73\times 10^{-1}$ & $6.63\times 10^{-1}$ & $2.27 \times 10^{-1}$ & $4.27\times 10^{-1}$ \\ FBPINN & $8.65\times 10^{-1}$ & $\mathbf{6.02\times 10^{-4}}$ & $1.57\times 10^{-1}$ & $2.08\times 10^{-1}$\\ AB-PINN (ours) & $\mathbf{5.47\times 10^{-3}}$ & $1.86\times 10^{-3}$ &$\mathbf{4.01\times 10^{-3}}$ & $\mathbf{1.67\times 10^{-2}}$ \\ \hline
\end{tabular}
    \captionsetup{width=.7\linewidth}
\caption{Comparison where the AB-PINN does not use residual-based subdomain addition and only exhibits local adaptivity of the subdomains.}
\label{tab:loc}
\end{subtable}

\begin{subtable}[t]{\textwidth}
\centering
\begin{tabular}{|c|c|c|c|}
\hline
 & Allen--Cahn equation & Forced Poisson equation & KdV equation \\
\hline
Standard PINN & $6.90\times 10^{-3}$ & $1.61\times 10^{-2}$ & $4.71\times 10^{-2}$ \\ 
AB-PINN+ (ours) & $\mathbf{2.46\times 10^{-4}}$ & $\mathbf{1.43\times 10^{-3}}$ & $\mathbf{1.55\times 10^{-2}}$  \\ 
\hline
\end{tabular}
    \captionsetup{width=.7\linewidth}
\caption{Comparison where new AB-PINN subdomains are added during training according to regions of high residual loss. }
\label{tab:glob}
\end{subtable}

\caption{Benchmarking AB-PINNs against the standard PINN architecture and FBPINNs. Table \ref{tab:loc} shows results for AB-PINNs with local adaptivity of the window functions, whereas the results reported in Table \ref{tab:glob} also consider the addition of new subdomains in regions of high PDE residual. Throughout, all models are trained 5 times with different random initializations and the errors for the model achieving the lowest PDE residual are reported.}
\label{tab:results}
\end{table}

Across each test, we use the Adam optimizer and decay all learning rates on an exponential schedule to $1\%$ of their initial value by the end of training \cite{kingma2014adam}. The initial learning rate for all MLP weights and biases is always set to $10^{-3}.$ The collocation points are always randomly uniformly resampled each iteration, and all global networks and subnetworks considered are multi-layer perceptrons (MLPs) with the hyperbolic tangent activation function. Unless otherwise specified, the reference window function considered is always the Gaussian $\psi(x) = e^{-x^2/2}$; see Remark~\ref{ex:1}. The number of training steps for the two training phases (see Section \ref{subsec:implementation}), number of collocation points, number of subdomains, initial learning rates for the window function parameters $\mu_i$ and $L_i$, and size of each MLP are problem-dependent.

 During optimization, the subnetwork parameters $\theta_i$, subdomain centers $\mu_i$, and tunable parameters of $L_i$ corresponding to each subdomain, as well as the global network parameters $\theta_0$, are all declared as separate parameter groups with their own optimizer and learning rates. When a new subdomain is added during training the associated tunable parameters are declared under a new optimizer, with learning rates that similarly decay on an exponential schedule, until all learnable window parameters are frozen at once for fine-tuning of the solution; see Section \ref{subsec:implementation}. Throughout training, the adaptive subdomain centers $\mu_i$ are restricted to remain inside the computational domain, and the diagonal of $L_i$ is constrained to be positive through the  absolute value function.

In many of our numerical experiments, we visualize the learned window functions $\phi_i$ in two-dimensions; see Figure~\ref{fig:adv} for example. Namely, we plot their centers $\mu_i$ with an \texttt{x} marker, we show a single thick contour line to indicate where the window function has reduced to half of its peak value, and we show a filled contour plot of the envelope $E(x) = \max_{1\leq i \leq N}\phi_i(x)$ of all window functions. As a convention for the visualizations, we assign $E(x) = 0$ for the standard PINN. For problems with periodic boundary conditions, we instead visualize the composition of the window functions with the Fourier coordinate embedding. Even if the window functions are Gaussian in the embedded coordinates, when visualized in the original coordinate system they appear distorted, and sometimes disconnected. Nevertheless, this visualization strategy reveals which regions of the spatio-temporal domain each subnetwork is responsible for.
  
The rest of this section is organized as follows. The experiments on local adaptivity of window functions, including the chirp waveform, the advection equation, and the Helmholtz equation are presented in Sections~\ref{subsec:chirp},~\ref{subsec:adv}, and~\ref{subsec:helm} respectively. Next, the tests on residual-based subdomain addition, including the Allen--Cahn equation, the forced Poisson equation, and the KdV equation are shown in Sections~\ref{subsec:ac}, \ref{subsec:poisson}, and \ref{subsec:kdv}. For each example comprehensive hyperparameter and optimization details are provided, along with plots of the learned solutions, PDE residuals, absolute errors, loss curves, and learned domain decompositions.

\subsection{Local Adaptivity of Subdomains}\label{subsec:local_adaptive_exp}
In the following experiments, we consider a basic AB-PINN architecture in which the number of subdomains is fixed throughout training. For every comparison with FBPINNs, we initialize the AB-PINN window functions in exactly the same configuration as the FBPINN model tested. Thus, the only difference between the FBPINN and AB-PINN models in this section is the adaptivity of the window functions and the inclusion of a global network.
\subsubsection{Chirp Waveform}\label{subsec:chirp}

We begin our numerical experiments with a motivating toy example which highlights the benefit of including adaptive basis functions within a PINN-based domain decomposition framework. In particular, we attempt to solve the first-order differential equation
\begin{equation}\label{eq:chirp}
    u'(x) = 2\pi \omega p \cos(2\pi \omega x^p) x^{p-1}, \qquad x\in (0,1), \qquad u(1) = 0,
\end{equation}
where $\omega,p \in \mathbb{N}$ are fixed. While the solution to \eqref{eq:chirp} is analytically known to be 
\begin{equation}\label{eq:chirp_sol}
    u(x) = \sin(2\pi \omega x^p), \qquad x\in (0,1).
\end{equation}
The problem \eqref{eq:chirp} provides a helpful example which can illustrate the difficulties of standard PINN frameworks and fixed domain decomposition strategies for solving multiscale oscillatory differential equations. For high values of $\omega$ the solution \eqref{eq:chirp_sol} becomes more oscillatory, while increasing the power $p$ localizes the oscillations. To produce a challenging test case, we select $\omega = 10$ and $ p = 10$. 
\begin{figure}[h!]
  \centering
  
  \subfloat[Training and testing loss history]
  {\includegraphics[width=0.32\textwidth]{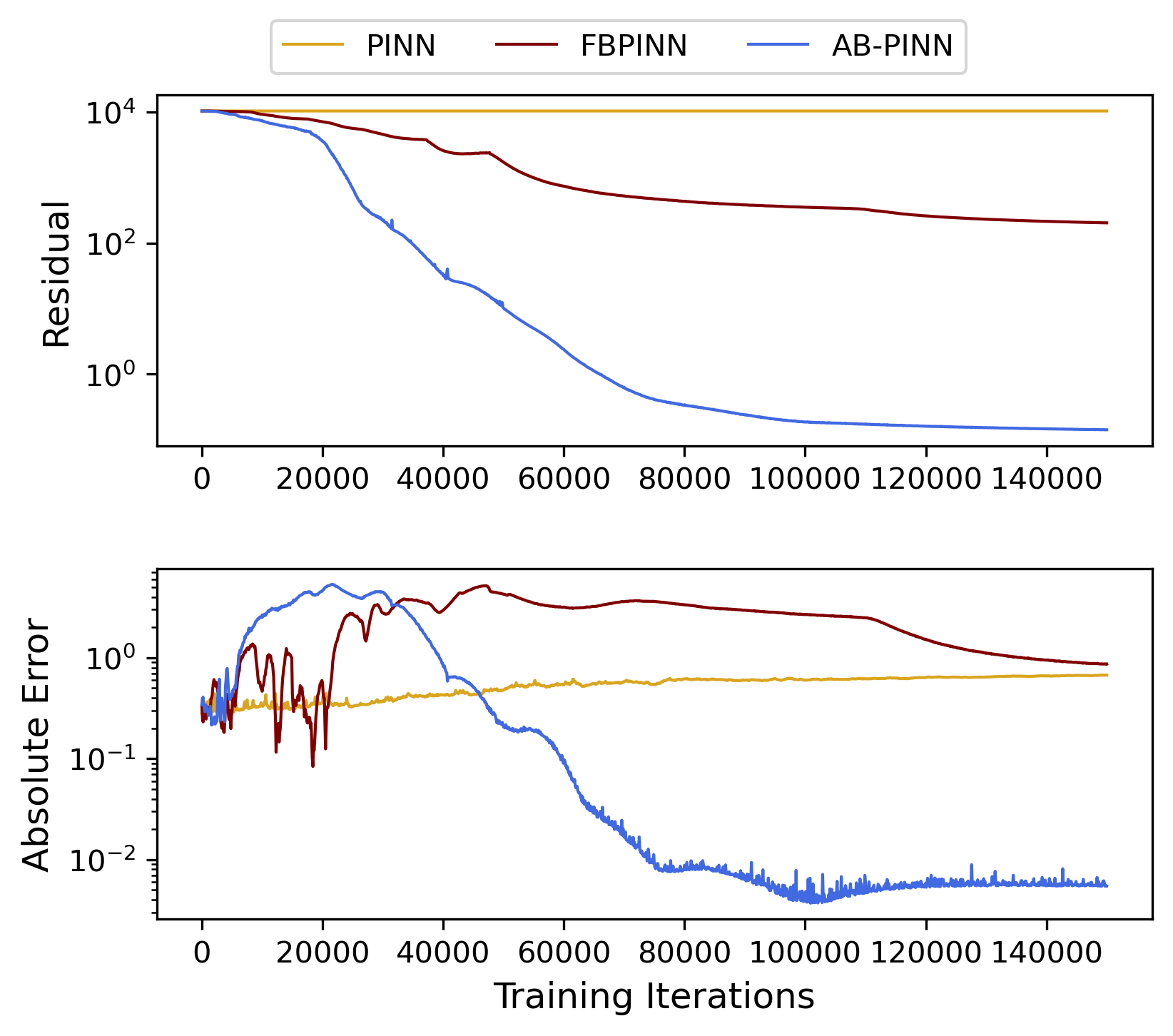}}
  \hfill
  \subfloat[Learned solutions]
  {\includegraphics[width=0.32\textwidth]{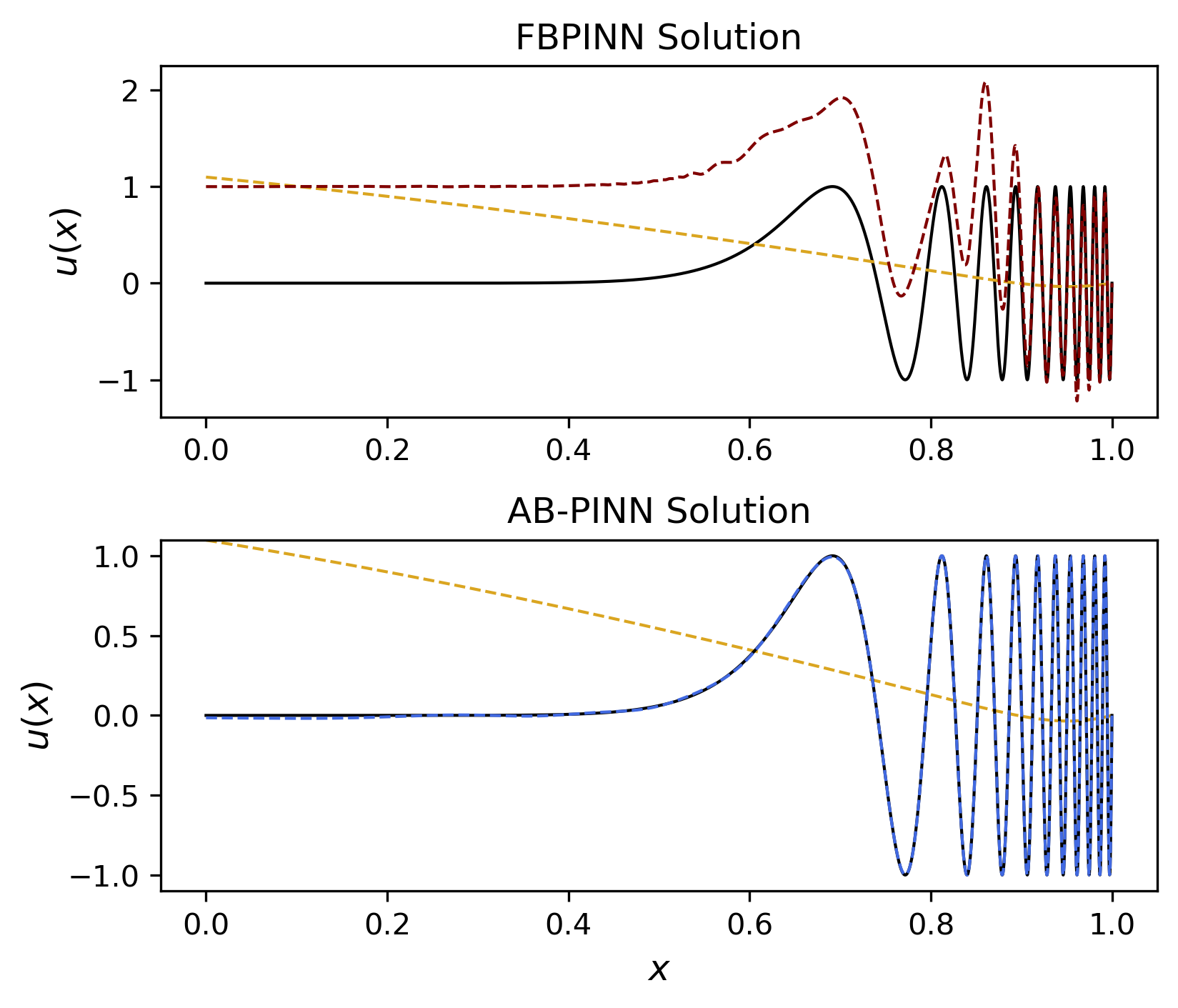} \label{fig:1b}}
  \hfill
  \subfloat[Window functions]
  {\includegraphics[width=0.32\textwidth]{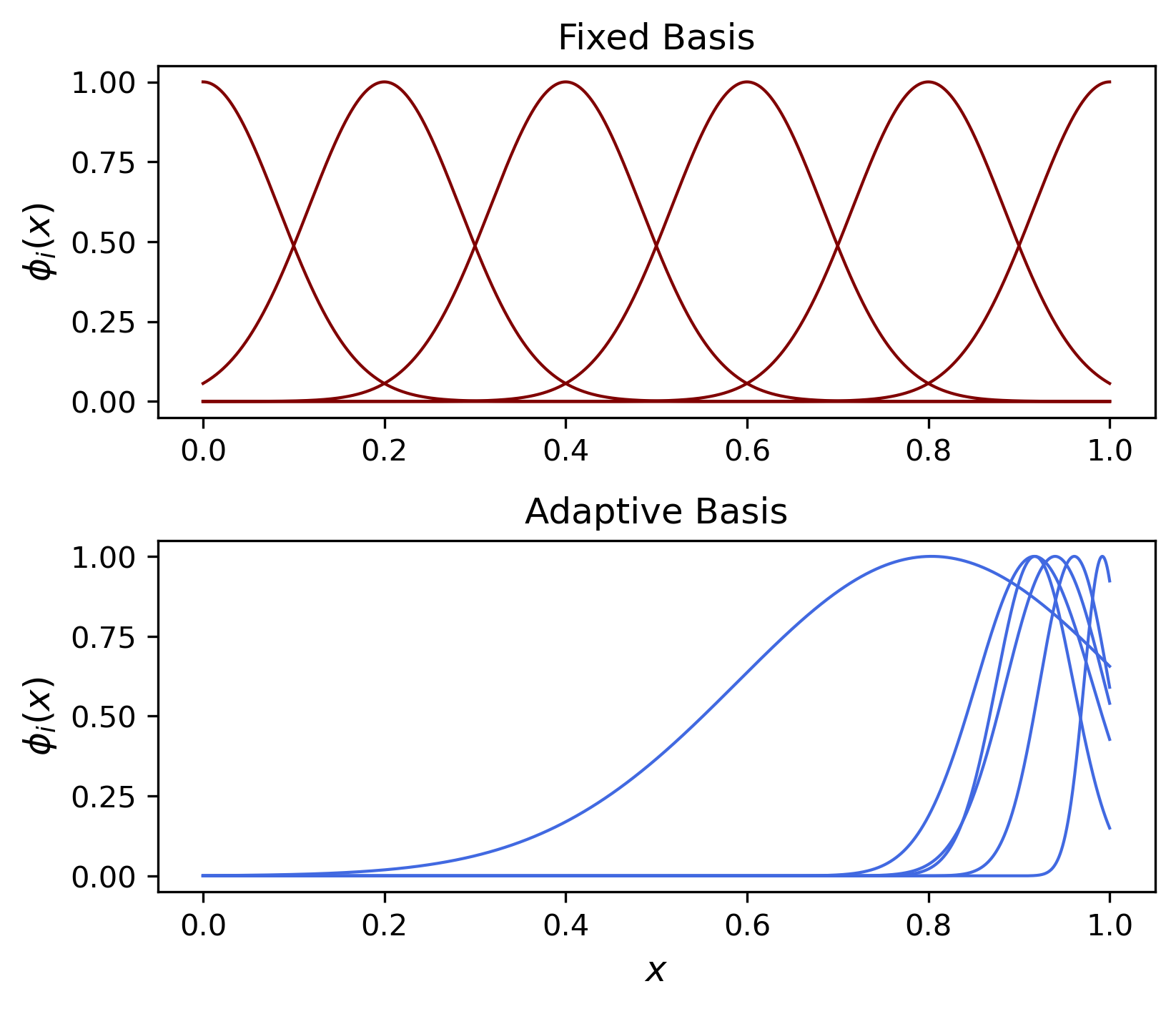}\label{fig:chirpc}}
  
  \caption{Solving a multiscale chirp waveform using a standard PINN, an FBPINN and an AB-PINN. In the second column the standard PINN solution is shown in yellow, while the FBPINN solution is plotted in red, the AB-PINN in blue, and the ground truth in black.}
  \label{fig:1}
\end{figure}

Figure \ref{fig:1} shows the results for solving \eqref{eq:chirp} using a standard PINN, an FBPINN and an AB-PINN. Across each test, the operator $$\Psi[u](x) = u(x) \tanh((x-1)/10)$$ is used to enforce $u(1) = 0$ as a hard constraint, and at each training iteration, $2\cdot 10^3$ collocation points are uniformly randomly sampled. The standard PINN is given by an MLP with two hidden layers, each consisting of $32$ nodes. Both the AB-PINN and FBPINN models use $N=6$ subdomains initialized according to \eqref{eq:coords} with $L_i = 6$ and $\mu_i$ linearly spaced over $[0,1]$; for a visualization see the top row of Figure \ref{fig:chirpc}. Each subnetwork in the AB-PINN model is an MLP with two hidden layers and 10 nodes in each layer, while the global network instead has 12 nodes in each layer. In the FBPINN model, each subnetwork has two hidden layers, each with 12 nodes per layer. We use a slightly different number of nodes per layer for the AB-PINN and FBPINN subnetworks to ensure a fair comparison, such that all models have roughly the same total number of tunable parameters. In this case, all models considered have roughly $10^3$ tunable parameters and are trained for $1.5\cdot 10^5$ iterations. We use learning rates of $10^{-3}$ and $10^{-2}$ for the learnable window function parameters $\mu_i$ and $L_i$, respectively; see \eqref{eq:coords}. After $5\cdot 10^4$ training iterations, the window function parameters $\mu_i$ and $L_i$ are frozen 
and no longer adapt.

As shown in Figure \ref{fig:1}, the standard PINN and FBPINN approaches struggle to represent the solution \eqref{eq:chirp_sol}, while the AB-PINN converges faster in terms of both the residual loss and the absolute solution error. The success of the AB-PINN is attributed to the flexibility of the adaptive window functions to concentrate on the region of the domain where the solution is most complex and the residual loss is highest. As shown in Figure \ref{fig:chirpc}, the window functions cluster nearby $x = 1$, where the solution is highly oscillatory. The corresponding subnetworks, whose inputs are normalized within the support of these basis functions, are then able to minimize the residual loss in this region with greater success than the FBPINN.

\begin{figure}[h!]
  \centering
  
  \subfloat[Training and testing loss history]
  {\includegraphics[width=0.32\textwidth]{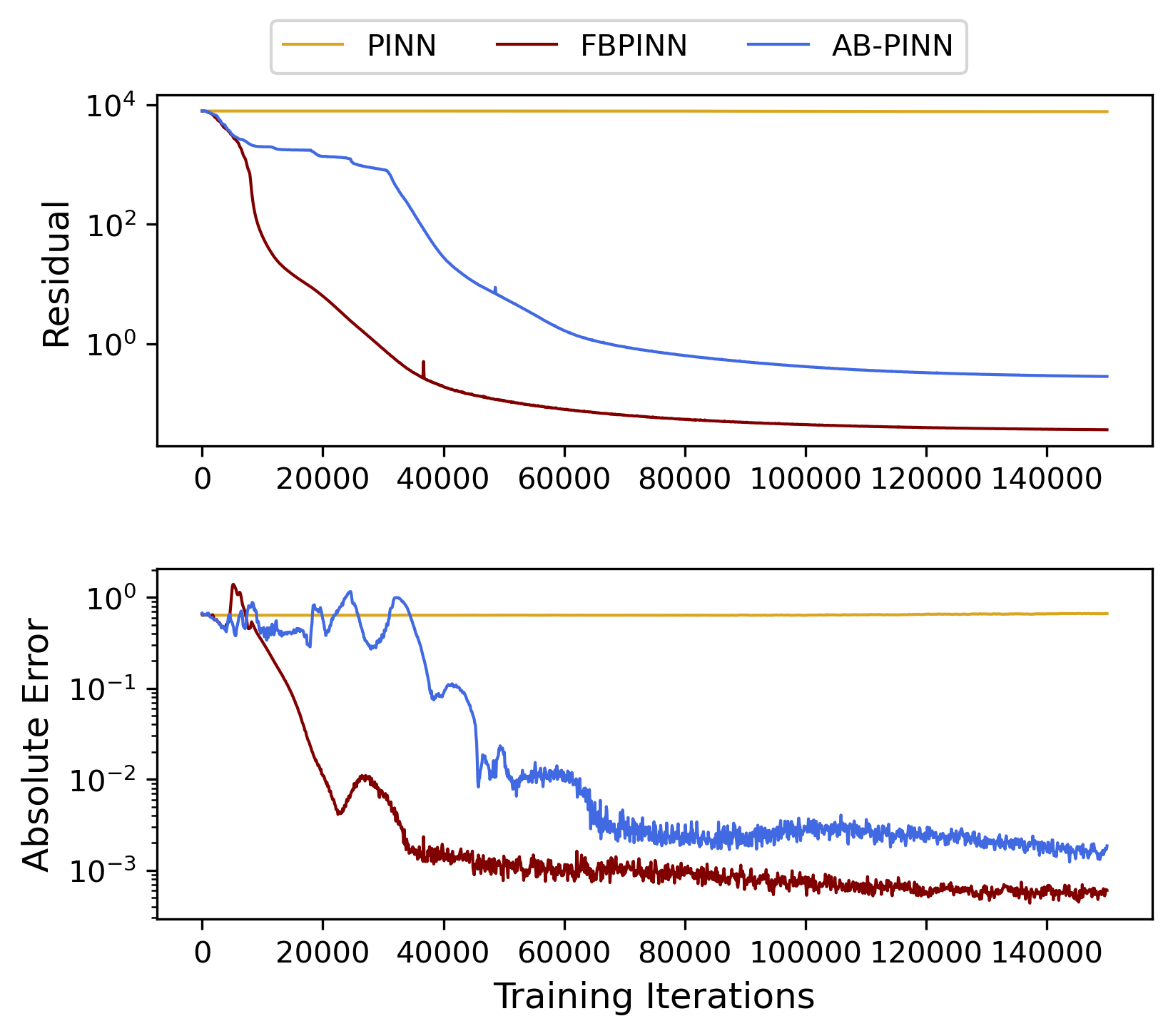}}
  \hfill
  \subfloat[Visualization of basis functions]
  {\includegraphics[width=0.32\textwidth]{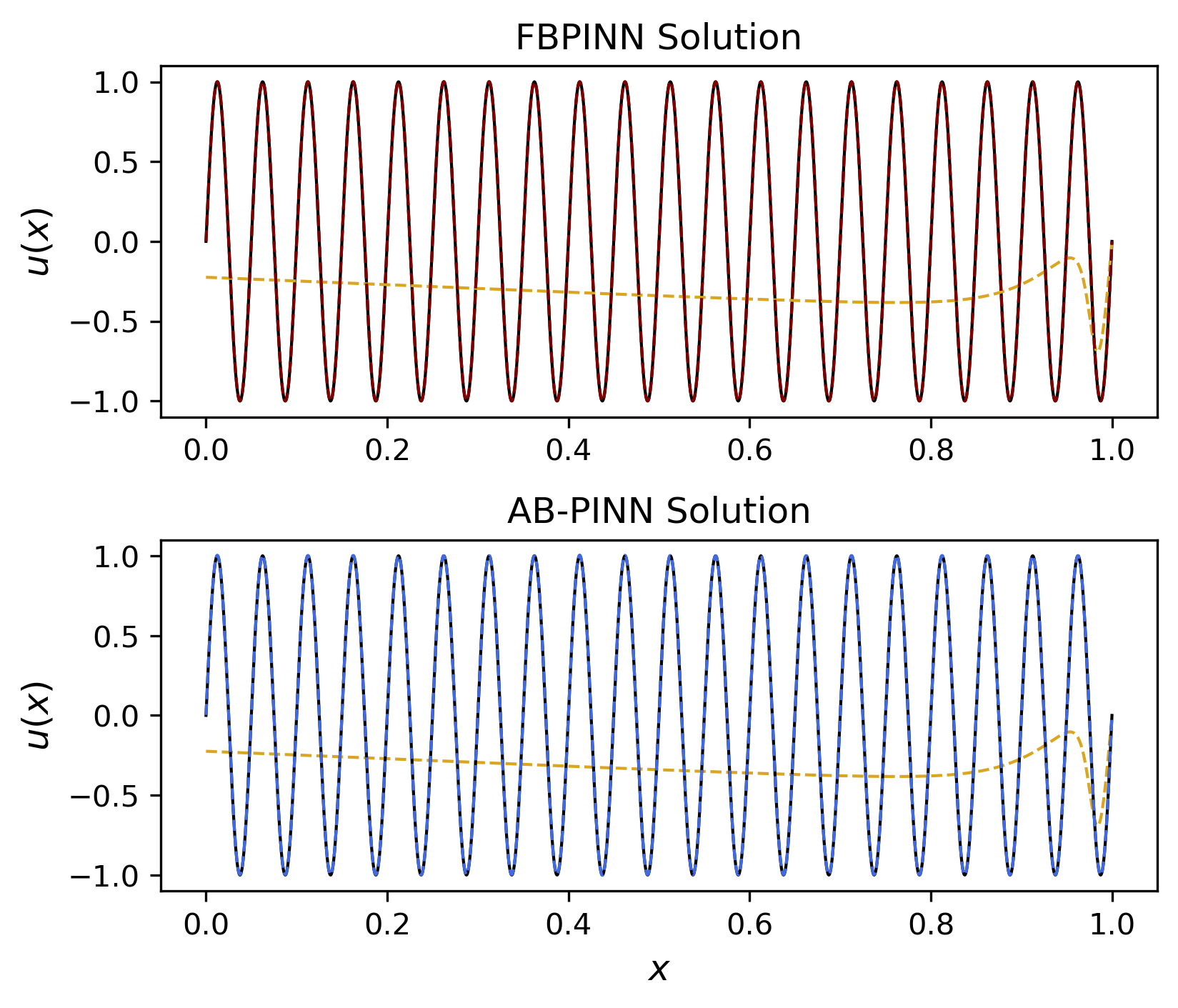} \label{fig:1b}}
  \hfill
  \subfloat[Visualization of learned solutions]
  {\includegraphics[width=0.32\textwidth]{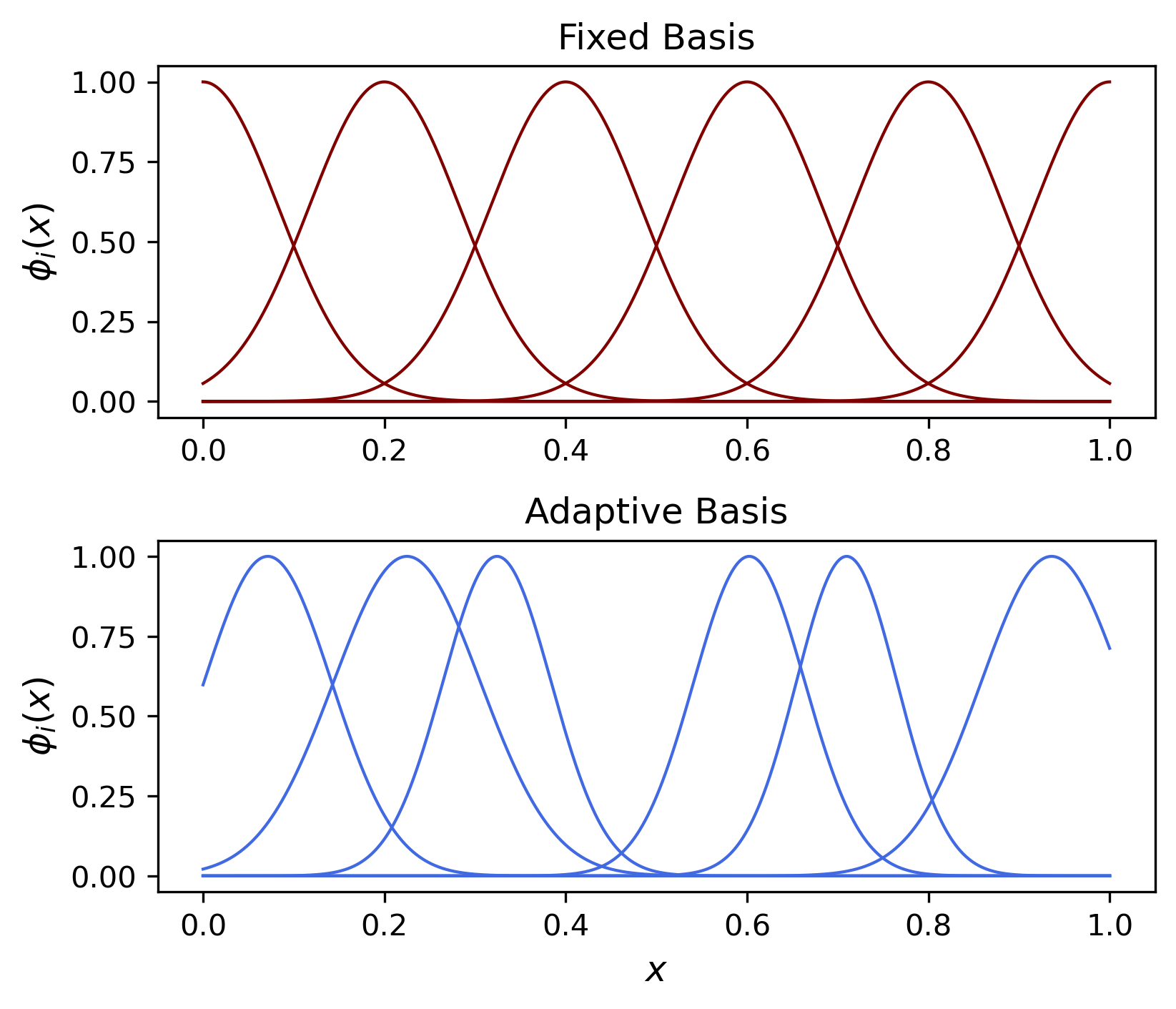}}
  
  \caption{Solving a uniform sine wave using a standard PINN, an FBPINN, and an AB-PINN. }
  \label{fig:sine}
\end{figure}

In Figure \ref{fig:sine}, we repeat the same experiment from Figure \ref{fig:1} with a differential equation that does not exhibit any multiscale behavior. We again consider \eqref{eq:chirp}, but now with $\omega = 20$ and $ p =1$. The reference solution for this equation is simply a sinusoidal oscillation with uniform frequency across the domain; see \eqref{eq:chirp_sol}. As shown in Figure \ref{fig:sine}, the AB-PINN architecture no longer has a significant advantage compared to the FBPINN, likely due to the fact that the uniform spacing of the FBPINN window functions is already optimal. The results of Figures \ref{fig:1} and \ref{fig:sine} collectively suggest that AB-PINNs are best suited for solving multiscale problems where the solution features are not uniformly distributed across the domain. 

\begin{figure}[h!]
    \centering   
    \subfloat[Loss curve comparison between AB-PINNs and FBPINNs for different choices of the reference window function $\psi$.]{\includegraphics[width=.8\textwidth]{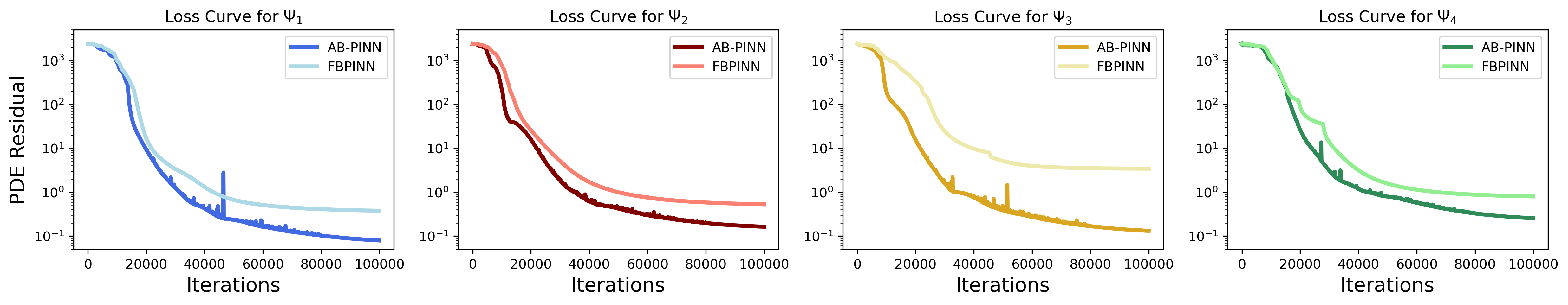} \label{fig:losseswindow}} \hfill
  \subfloat[Visualization of the FBPINN basis (top), the learned AB-PINN basis (middle), and the learned AB-PINN solution (bottom)]{\includegraphics[width=\textwidth]{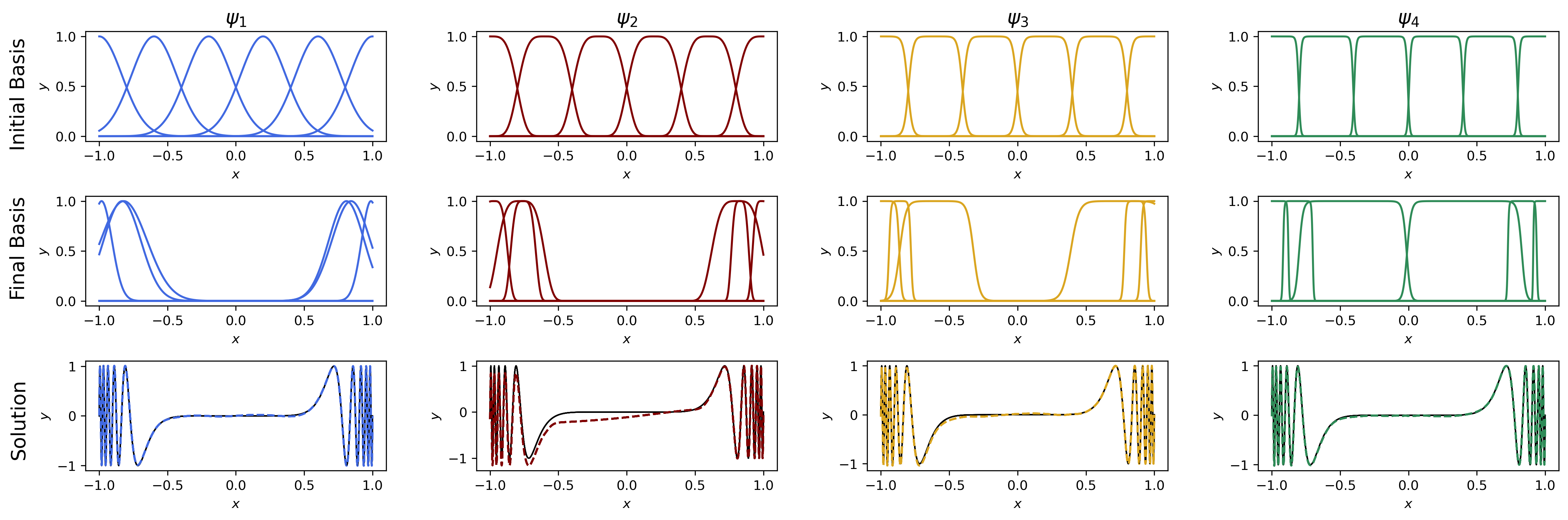}\label{fig:basis}}
   \caption{Studying the effect of the choice of reference window function on the performance of AB-PINNs for a multiscale chirp problem.}\label{fig:basis_fig} 
\end{figure}

In Figure \ref{fig:basis_fig}, we study the effect of the choice of the reference window function $\psi$ (see \eqref{eq:coords}) for solving \eqref{eq:chirp}. For this experiment, we consider \eqref{eq:chirp} with $x\in (-1,1)$, $\omega = 5$, and $p = 9$. We consider the following reference window functions for defining both our AB-PINN and FBPINN models:
\begin{align}\label{eq:window_choices}
\begin{cases}
  \displaystyle   \psi_1(x) = e^{-x^2/2}  \\
  \displaystyle    \psi_2(x)= e^{-x^4/(4 \ln 2)}\\
  \displaystyle    \psi_3(x)= \frac{1}{1+e^{-10(\sqrt{2\ln 2}+x)}}\cdot\frac{1}{1+e^{-10(\sqrt{2\ln 2}-x)}},\\
  \displaystyle    \psi_4(x)= \frac{1}{1+e^{-10(2\ln 2 -x^2)}}.
\end{cases}
\end{align}
The functions in \eqref{eq:window_choices} are designed such that each reaches a maximum of approximately $1$ at $x = 0$ 
and symmetrically reduce to $1/2$ at $x=\pm \sqrt{2 \ln 2}$. We remark that $\psi_3$ belongs to the collection of window functions considered in the original FBPINN paper \cite{moseley2020solving}. 

For the study in Figure \ref{fig:basis_fig}, all initial learning rates are set to $10^{-3}.$ The individual window functions $\phi_i$ are initialized according to \eqref{eq:coords} with $L_i = 6$ and $\mu_i$ linearly spaced over $[-1,1]$; for a visualization see the top row of Figure \ref{fig:basis_fig}. The loss curves depicted in Figure \ref{fig:losseswindow} indicate that for all choices of the reference window function, an AB-PINN framework in which the window function parameters adapt outperforms the static FBPINN model. The middle row of Figure \ref{fig:basis_fig} shows the learned window functions after their parameters have been frozen following $8\cdot 10^4$ training iterations, and the predicted AB-PINN solution is shown in Figure \ref{fig:basis_fig}. We observe that the window functions with flat peaks and quickly decaying tails, e.g. $\psi_4$, do not display as much adaptivity as those without flat peaks and with slowly decaying tails, e.g. $\psi_1$. In the remainder of our numerical experiments we select $\psi_1$ as our reference window function.

\subsubsection{Advection Equation}\label{subsec:adv}

\begin{figure}[h!]
  \centering
  
  \subfloat[Convergence of the PDE residual loss and the absolute error of the predicted solution]
  {\includegraphics[width=.7\textwidth]{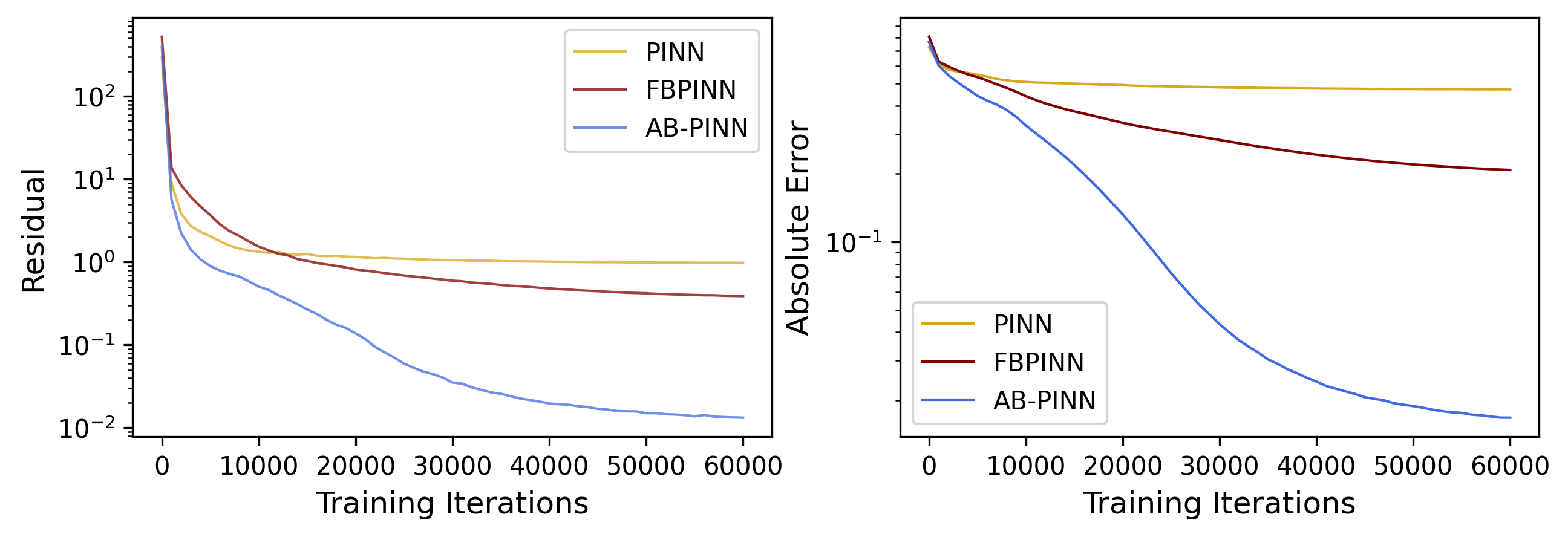}}
  \\
  \subfloat[Visualization of the learned basis functions, PINN solution, PDE residual, and absolute solution error.]
  {\includegraphics[width=.9\textwidth]{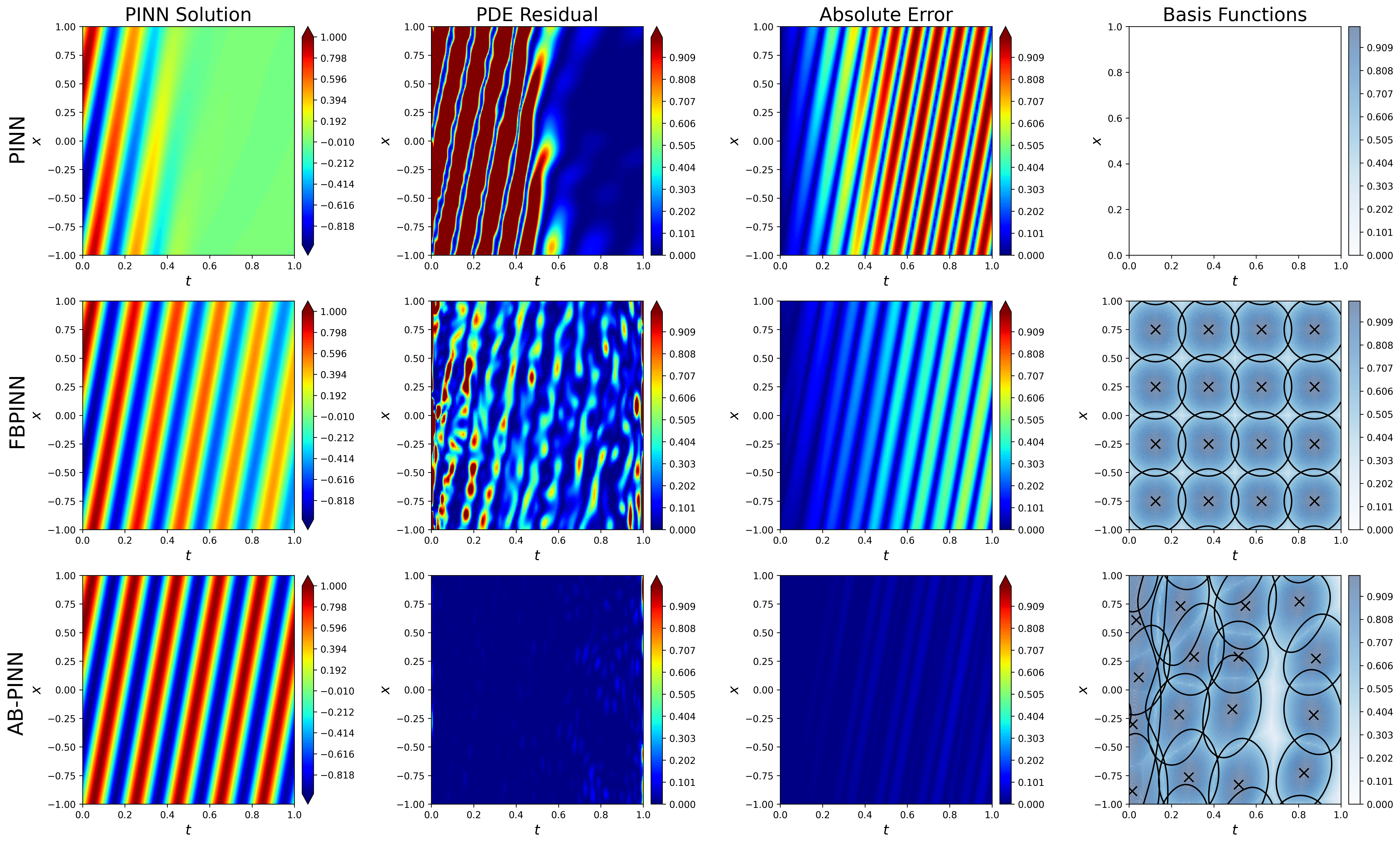} \label{fig:adv2}}

  \caption{Solving the advection equation using PINNs, FBPINNs, and AB-PINNs.}
  \label{fig:adv}
\end{figure}

We next consider the advection equation
\begin{align}
    \partial_t u(t,x) + c \cdot\partial_x u(t,x) = 0,& \qquad t\in (0,1) ,\, x\in (-1,1)  \label{eq:advection1}\\
    u(0,x) = \sin(\pi x),& \qquad x\in (-1,1) \label{eq:advection2}\\
    u(t,1)  = u(t,-1),& \qquad t\in [0,1], \label{eq:advection3}
\end{align}
with $c= 10.$ The ground truth solution to \eqref{eq:advection1}-\eqref{eq:advection3} is analytically given by 
$u(t,x) = \sin(\pi(x-ct))$, which is visualized in Figure \ref{fig:advref}. In Figure \ref{fig:adv}, we approximate the solution of the advection equation using standard PINNs, FBPINNs, and AB-PINNs. All models considered use the constraining operator 
$$\Psi[u](t,x) = u(t,x)\tanh(t)+\sin(\pi x)$$
to enforce the initial condition \eqref{eq:advection2} as a hard constraint, while input-coordinate Fourier embeddings are used to enforce the periodic boundary condition \eqref{eq:advection3}; see Section \ref{subsec:periodicity}. 

The FBPINN and AB-PINN models both use $N=16$ subdomains, with the AB-PINN subdomains initialized according to the FBPINN layout shown in Figure \ref{fig:adv}. Each MLP used in this test has four hidden layers, with the number of nodes in each layer carefully selected such that the PINN, FBPINN, and AB-PINN models all have roughly $7\cdot 10^3$ tunable parameters. The initial learning rates for the tunable window function parameters $\mu_i$ and $L_i$ are $5\cdot 10^{-4}$ and $10^{-3}$, respectively. Each model is trained for $6\cdot 10^4$ iterations, with $1000$ collocation points uniformly sampled at each step. After $10^4$ training steps, the tunable window function parameters in the AB-PINN model are frozen.

As shown in Figure \ref{fig:adv}, the AB-PINN converges faster and with higher accuracy than both the standard PINN and the FBPINN. Several notable features of the learned domain decomposition shown in the bottom row of Figure \ref{fig:adv} enable this faster convergence. First, the window functions adapt to the geometry of the underlying solution, stretching along its characteristic lines. Moreover, we observe that the spatial distribution of window functions is skewed towards $t = 0$, indicating that the adaptive subdomains place a higher emphasis on learning the solution at earlier times. Such behavior is similar to curriculum training and causality-inducing reweighting or resampling strategies for PINNs \cite{wight2021solving,krishnapriyan2021characterizing,daw2023mitigating,wang2024respecting}, all of which encourage the solution to be learned sequentially along time for more accurate results. We note that the AB-PINN architecture should be viewed as an enhancement rather than a replacement of these approaches.

\subsubsection{Helmholtz Equation}\label{subsec:helm}
\begin{figure}[h!]
  \centering
  
  \subfloat[Training and testing loss history]
  {\includegraphics[width=.8\textwidth]{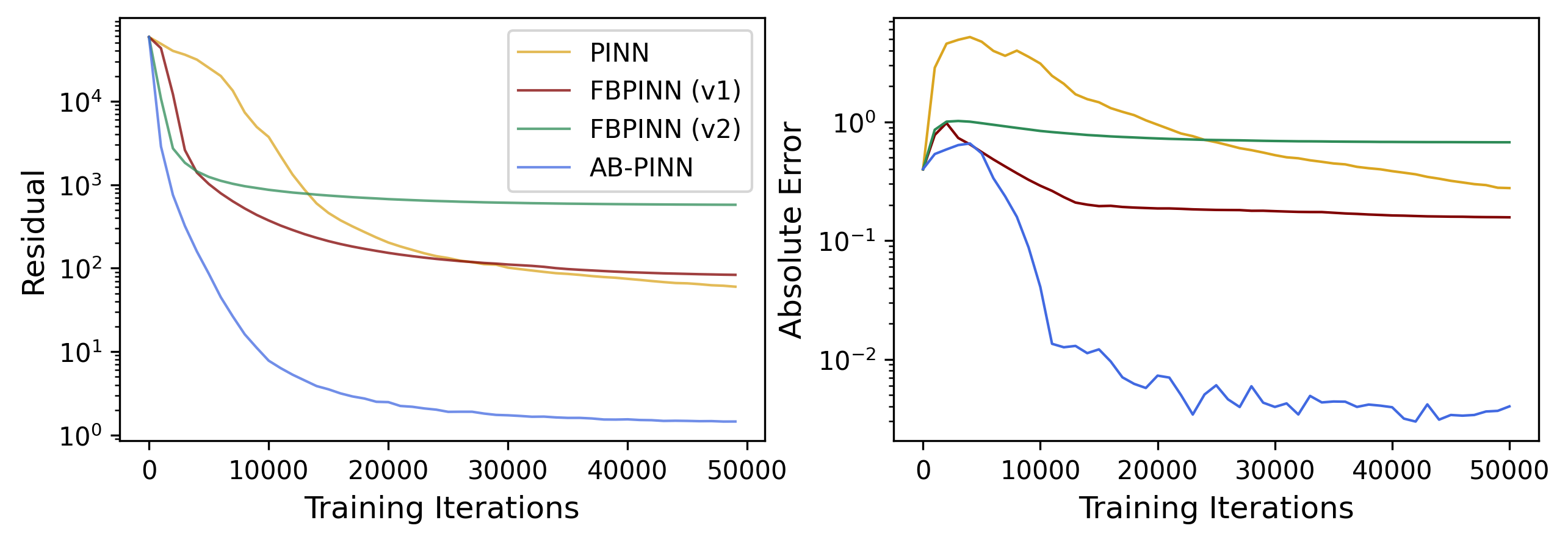}}
 \\
  \subfloat[Visualization of the PINN solution, PDE residual, and absolute solution error after training both the fixed basis and adaptive basis approaches.]
  {\includegraphics[width=\textwidth]{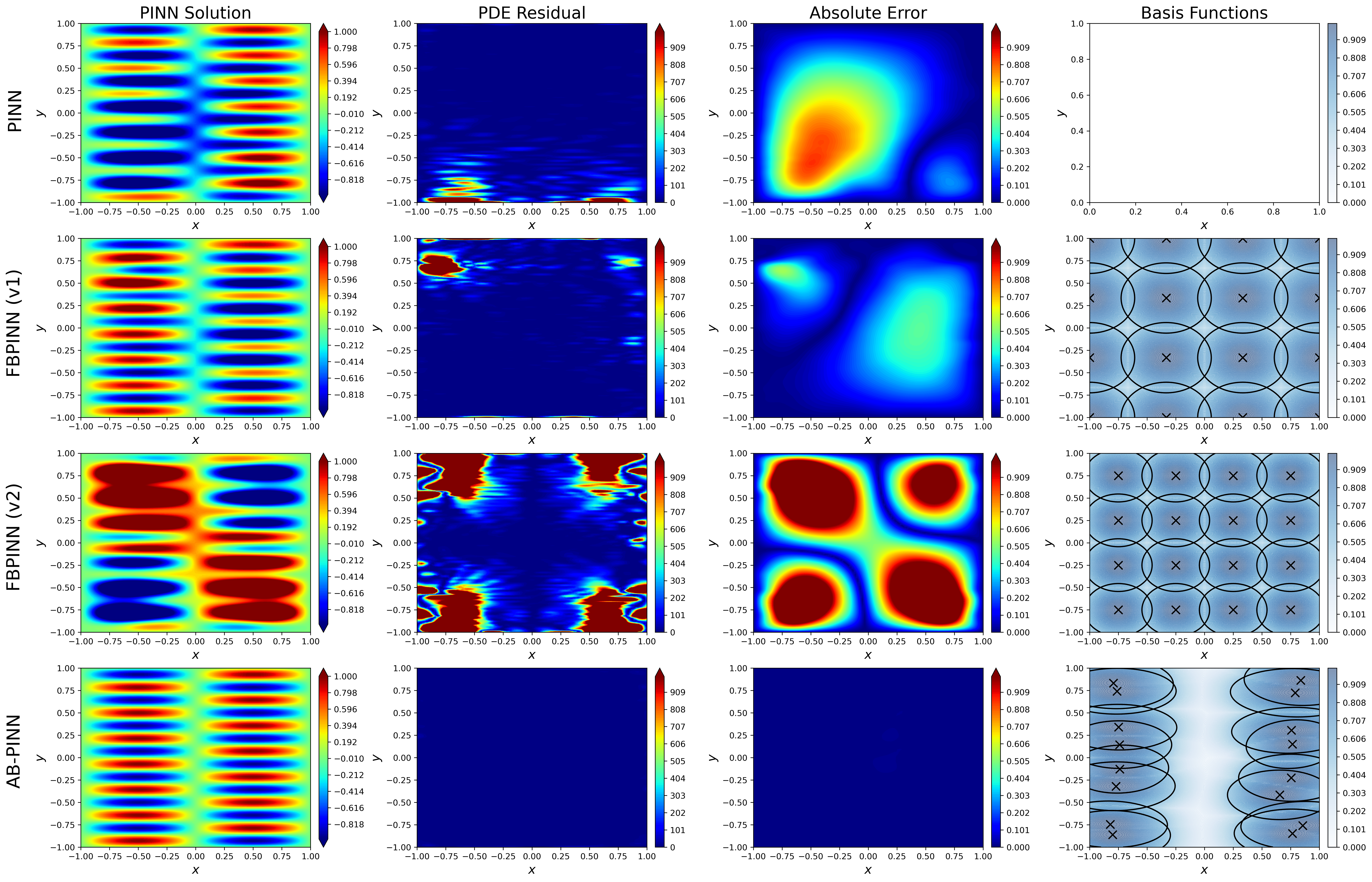} \label{fig:1b}}

  \caption{Solving the Helmholtz equation \eqref{eq:helmholtz1}-\eqref{eq:helmholtz2} using a vanilla PINN, a fixed basis approach, and an adaptive basis approach. The adaptive basis fits the multiscale geometry of the PDE and leads to faster convergence of the PDE residual loss and a more accurate solution.}
  \label{fig:helmholtz}
\end{figure}

We now consider the two-dimensional Helmholtz equation 
\begin{align}
 \Delta u (x,y)+ k^2 u(x,y) &= (k^2 - (a\pi)^2 - (b\pi)^2) \sin(a\pi x)\sin(b\pi y), \qquad (x,y)\in (-1,1)^2, \label{eq:helmholtz1}\\
  u(x,\pm 1) &= 0,\qquad u(\pm 1, y) = 0, \qquad x\in [0,1],\, y\in [0,1]\label{eq:helmholtz2}
\end{align}
where we set $ k = 2$, $a = 1$, and $ b = 7$. The solution of \eqref{eq:helmholtz1}-\eqref{eq:helmholtz2} is analytically given by 
\begin{equation}\label{eq:sol_helmholtz}
    u(x,y) = \sin(a\pi x)\sin(b\pi y),
\end{equation}
which is visualized in Figure \ref{fig:helmref}. By selecting $a = 1$ and $ b = 7$ we produce a solution with rapid oscillations in one direction and slow oscillations in the other. Without this prior knowledge, it is reasonable to initialize the window functions to uniformly cover the domain on a fixed isotropic grid.  In Figure \ref{fig:helmholtz}, we show that allowing the window functions to evolve during training under our AB-PINN framework lets them adapt to the geometry of the underlying PDE solution, thereby leading to a model which produces more accurate predictions.

In particular, we show the results of training a standard PINN, an FBPINN, and an AB-PINN to approximate the solution of \eqref{eq:helmholtz1}-\eqref{eq:helmholtz2}. Throughout, we use the constraining operator
\begin{equation}\label{eq:constrainH}
    \Psi[u](x,y) = u(x,y) \tanh(x-1)\tanh(x+1)\tanh(y-1)\tanh(y+1)
\end{equation}
to enforce the boundary condition \eqref{eq:helmholtz2}. Both the FBPINN and AB-PINN models use $N = 16$ subdomains. We test two different subdomain layouts for the FBPINN model (see Figure \ref{fig:helmholtz}), neither of which is able to match the accuracy of the AB-PINN. The AB-PINN model is initialized such that the window functions are in the same configuration as the second FBPINN model tested; see Figure \ref{fig:helmholtz}. All models considered use global networks and subnetworks with two hidden layers with a variable number of nodes per model, such that each model has roughly $10^4$ free parameters. In particular, the standard PINN uses four hidden layers with 57 nodes each, the FBPINNs use two hidden layer subnetworks with 23 nodes in each layer, and the AB-PINN uses two hidden layer subnetworks with 22 nodes per layer along with a two hidden layer global network with 20 nodes in each layer. We train each model for $5\cdot 10^4$ training iterations, uniformly sampling $10^3$ collocation points at each step. All initial learning rates are set to $10^{-3}$, aside from those for the window parameters $L_i$ which we set as $5\cdot 10^{-4}$. After the first $10^4$ iterations, the tunable window function parameters in the AB-PINN model are frozen. While in Figure \ref{fig:helmholtz}, we use solely the local adaptivity of the window functions to learn an accurate AB-PINN model for the Helmholtz equation, in the following section, we show how the residual-based subdomain addition strategy introduced in Section \ref{subsec:addition} can be used to solve the Helmholtz problem as well; see Figure \ref{fig:helmholtz_add}.

\subsection{Residual-Based Subdomain Addition}\label{subsec:global_adaptive_exp}
In Section \ref{subsec:local_adaptive_exp} we performed several comparisons between AB-PINNs, FBPINNs, and standard PINNs in the case where the number of AB-PINN window functions is fixed throughout training. In this section, we study the residual-based subdomain addition technique introduced in Section \ref{subsec:addition}. Our experiments show that adding new AB-PINN subdomains during training can be useful for avoiding unwanted local minima and delivering needed expressive power in regions of the domain where the PINN is struggling to converge. As a reminder, we write AB-PINN+ as shorthand to denote an AB-PINN model whose subdomains were introduced throughout training using residual-based subdomain addition.

\subsubsection{Helmholtz Equation}
\begin{figure}[h!]
    \centering   
    \subfloat[Training and testing loss history with and without residual-based subdomain addition. We write AB-PINN+ to denote the model which uses residual-based subdomain addition. The red \texttt{x} indicates when new subdomains are added.]{\includegraphics[width=.8\textwidth]{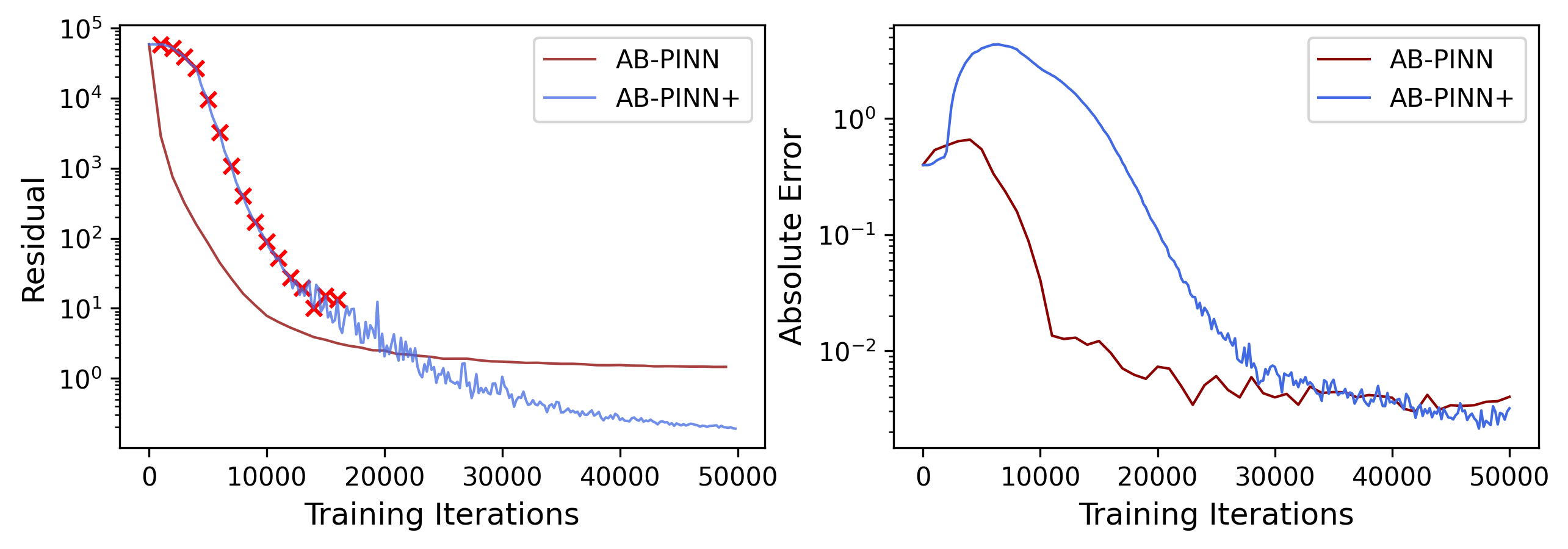} \label{fig:h_add1}} \hfill
  \subfloat[Visualizing the AB-PINN+ progress as training progresses. Each column shows the solution, residual, and learned domain decomposition after a fixed number of iterations have passed. The red \texttt{x} marker indicates the newly introduced subdomain.]{\includegraphics[width=\textwidth]{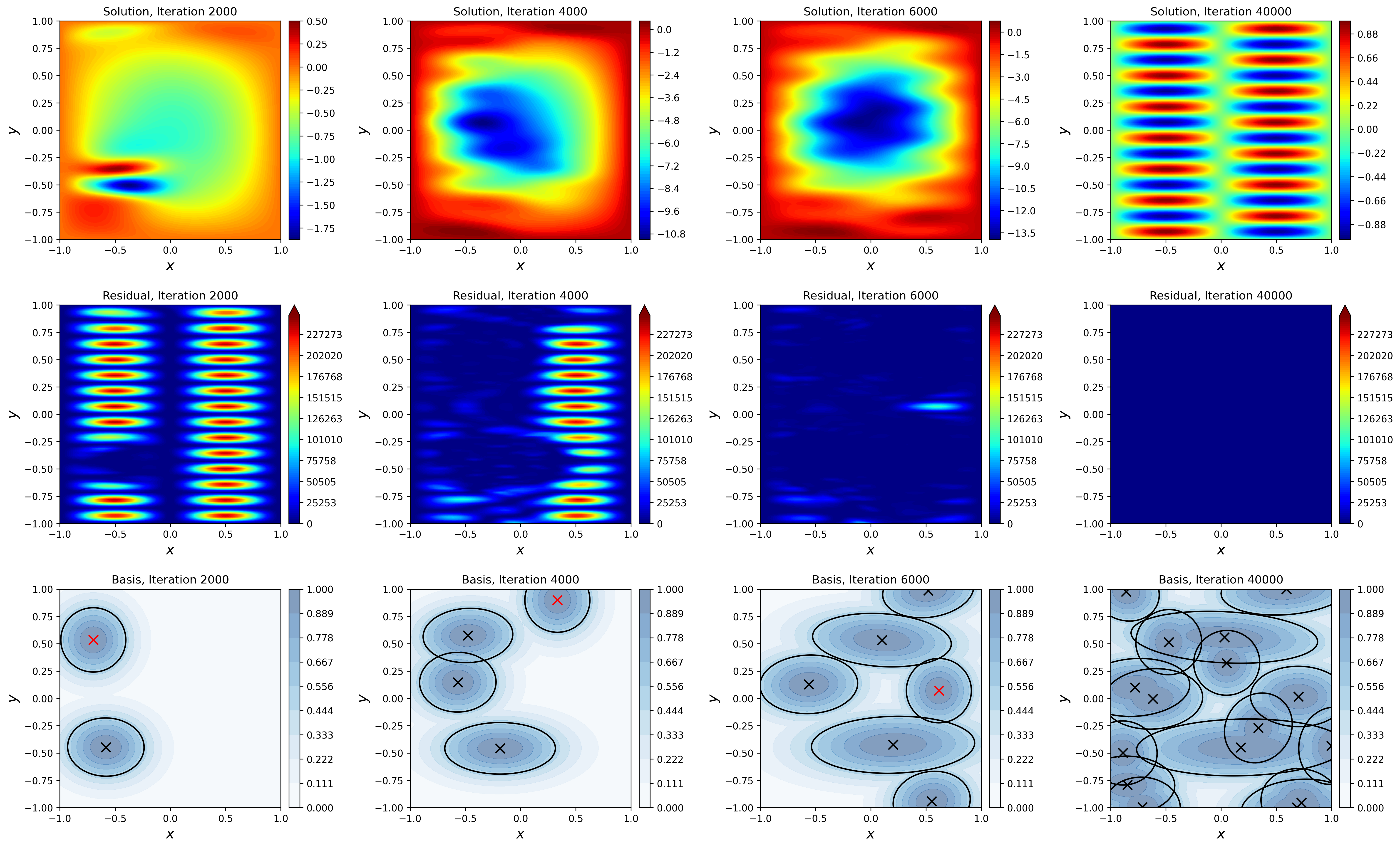}\label{fig:h_add2}}
   \caption{Solving the Helmholtz equation using the AB-PINN+ framework. Figure \ref{fig:h_add1} compares its convergence with the AB-PINN from section \ref{subsec:helm}, and Figure \ref{fig:h_add2} visualizes the training progress. }\label{fig:helmholtz_add} 
\end{figure}

We begin by illustrating how the residual-based subdomain addition can be also used to effectively solve the Helmholtz problem introduced in Section \ref{subsec:helm}. 
In Figure~\ref{fig:helmholtz_add}, we solve the Helmholtz equation \eqref{eq:helmholtz1}-\eqref{eq:helmholtz2} using the AB-PINN+ framework, progressively adding in more subdomains throughout training. Figure \ref{fig:h_add1} compares the AB-PINN+ performance to the fixed AB-PINN initialization from Section \ref{subsec:helm}. Each column of Figure \ref{fig:h_add2} then shows the predicted solution, PDE residual, and AB-PINN+ subdomains after a fixed number of iterations have elapsed. 

We begin the experiment with a single two-hidden layer global network with 20 nodes per layer. Every $10^3$ steps of training we add a new window function and corresponding subnetwork with two hidden layers each with 22 nodes per layer. As described in Section \ref{subsec:addition}, the location of the new subdomain is adaptively determined by sampling from a density which is proportional to the current PDE residual. After the subdomains are introduced, using the same initialization as in Section \ref{subsec:helm}, i.e., $L_i = 3I$ where $I$ is the identity matrix, their tunable parameters $\mu_i$ and $L_i$ are free to evolve dynamically until $2.5\cdot 10^4$ iterations have elapsed, at which point their learning rates are set to zero. As in Section \ref{subsec:helm}, the initial learning rates of the subdomain parameters $\mu_i$ and $L_i$ are $10^{-3}$ and $5\cdot 10^{-4}$, respectively. In total, we introduce $N = 16$ subdomains and arrive at a final AB-PINN+ architecture with approximately $10^4$ free parameters, same as the fixed AB-PINN from Section \ref{subsec:helm}.

As shown in Figure \ref{fig:h_add2}, the PDE residual reduces significantly in the regions where new subdomains have been added. This suggests that the addition of new AB-PINN+ subdomains can be used to reduce localized regions of high PDE residual. Notably, while the FBPINN and PINN architectures studied in Section \ref{subsec:helm} were not able to provide an accurate solution to the Helmholtz equation, the AB-PINN+ is successful despite having roughly the same number of tunable parameters. Finally, we note that the learned domain decomposition from residual-based subdomain addition (AB-PINN+) differs significantly from the learned domain decomposition when all basis functions are initialized at the start of training (AB-PINN); see Figures \ref{fig:helmholtz} and \ref{fig:helmholtz_add}. In the next section, we provide a detailed comparison of the learned domain decomposition under these two setups for a different PDE. 

\subsubsection{Locally Forced Poisson equation}\label{subsec:poisson}

In this example, we demonstrate that the AB-PINN+ domain decompositions arrived at via residual-based subdomain addition can, in certain situations, yield a more accurate solution than the AB-PINN decomposition learned using solely the local adaptivity of the window functions. In particular, we consider the locally forced Poisson equation
\begin{align}
    \Delta u(x,y) &= \sum_{i=1}^9 \frac{(x-c_{i,1})^2+(y-c_{i,2})^2-2\sigma^2}{\sigma^4}\exp\Bigg(- \frac{(x-c_{i,1})^2+(y-c_{i,2})^2}{2\sigma^2}\Bigg),\quad (x,y)\in \Omega\label{eq:poisson1}\\
    u(x,y)&= \sum_{i=1}^9 \exp\Bigg(- \frac{(x-c_{i,1})^2+(y-c_{i,2})^2}{2\sigma^2}\Bigg),\qquad (x,y)\in \partial \Omega, \label{eq:poisson2}
\end{align}
where  $\Omega = [-1,1]^2$, $\sigma = 0.025$, and $c_{i,j}  \in (-1,1)$. The solution to \eqref{eq:poisson1}-\eqref{eq:poisson2} over $\Omega$ is given by the sum of Gaussians used to specify the boundary condition in \eqref{eq:poisson2}; see Figure \ref{fig:poissref}. For small values of $\sigma$, the boundary condition \eqref{eq:poisson2} can be approximated as $u(x,y) \approx 0$ for $(x,y)\in \partial \Omega$, and thus we use the same constraining operator \eqref{eq:constrainH} from Section \ref{subsec:helm} to enforce this approximate boundary condition.

\begin{figure}[h!]
  \centering
  
  \subfloat[PDE residual loss and absolute error of the predicted solutions. The red \texttt{x} in the loss curve indicates when a new subdomain is added.]
  {\includegraphics[width=.8\textwidth]{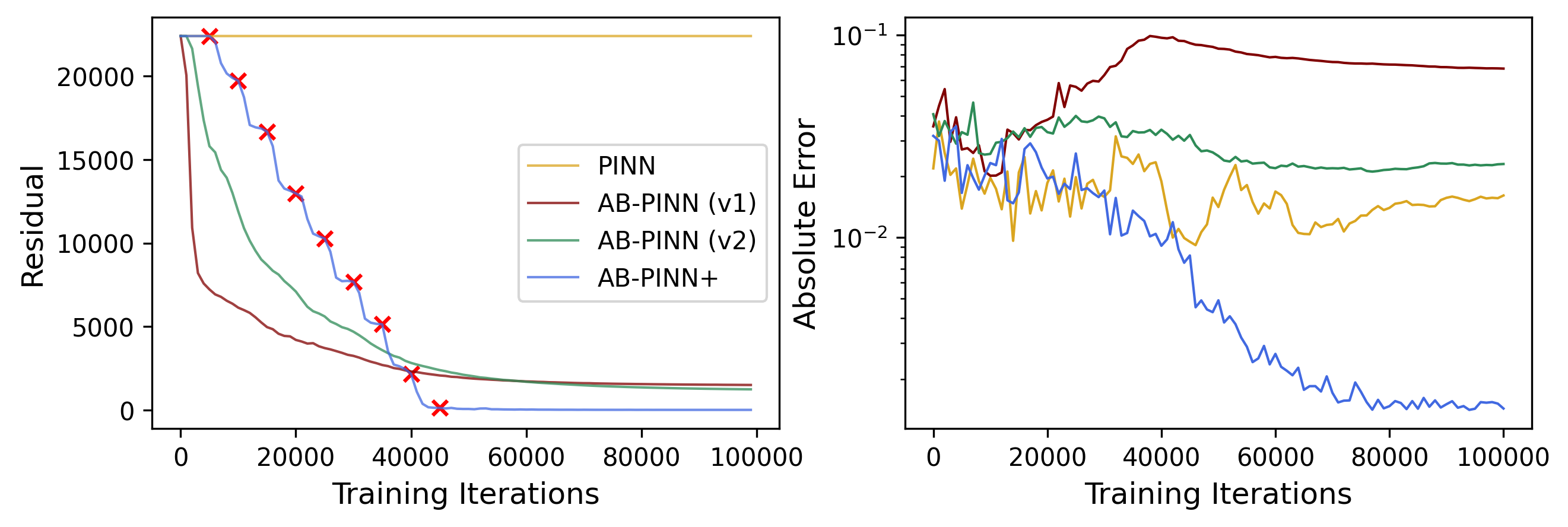}}
 \\
  \subfloat[Visualization of the PINN solution, PDE residual, and absolute solution error, and learned domain decomposition.]
  {\includegraphics[width=.9\textwidth]{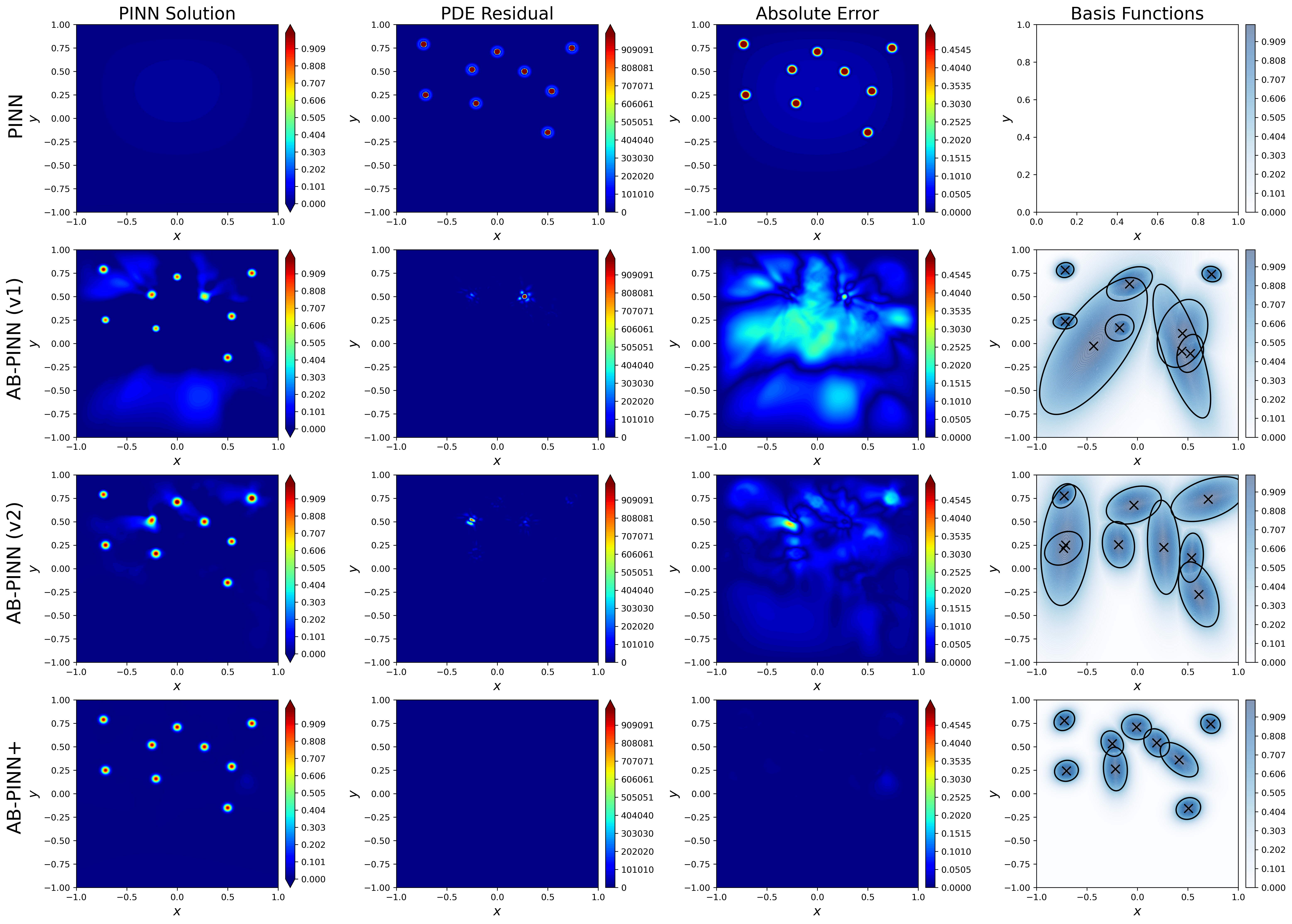} \label{fig:poisson2}}

  \caption{Solving the locally forced Poisson equation using the AB-PINN+ (bottom row of Figure \ref{fig:poisson2}) leads to better accuracy than relying solely on local adaptivity of the window functions initialized from a uniform configuration (middle two rows of Figure \ref{fig:poisson2}).}
  \label{fig:poisson}
\end{figure}
In Figure \ref{fig:poisson}, we solve the locally forced Poisson equation using a standard PINN, two AB-PINN models where all subdomains are initialized at the beginning of training, as well as an AB-PINN+ which makes use of residual-based subdomain addition. Throughout, all models tested have roughly $5\cdot 10^3$ tunable parameters, are trained for $10^5$ iterations, and use $1000$ uniform random collocation points per iteration. In particular, the PINN uses two hidden layers with 70 nodes each, while the AB-PINN and AB-PINN+ models use a two hidden layer networks with 20 nodes per layer for both the global network and subnetworks.  The learning rates of the tunable window parameters $\mu_i$ and $L_i$ are initialized at $10^{-3}$ and $5\cdot 10^{-3}$, respectively. All AB-PINN models use exactly $N = 9 $ subdomains and freeze the tunable parameters of the window functions after $7\cdot 10^4$ steps. For the AB-PINN+, we begin with a single global network and add in a new subdomain every $5\cdot 10^3$ training iterations, until $N = 9$ subdomains have been introduced. The two AB-PINN models based on local adaptivity are initialized with the centers $\mu_i$ uniformly spaced over the unit square, while the parameters $L_i$ are initialized as $5 I$ and $2.75 I$, respectively, where $I$ denotes the identity matrix. For the AB-PINN+, we set $L_i = 5 I.$ 

If the AB-PINN models configures their subdomains optimally, then one window function will be concentrated on each peak of the solution. In Figure \ref{fig:poisson}, we observe that the AB-PINN+ with residual-based subdomain addition indeed achieves this goal and leads to an accurate solution, while the two AB-PINNs with local adaptivity starting from a square grid layout of window functions struggle. Indeed, each time a new AB-PINN+ subdomain is introduced it targets a region of high PDE residual, which for this problem corresponds to one of the Gaussian peaks. After the subdomain is introduced, the residual in the targeted region decreases, and thus the next subdomain is placed on a different peak. This leads to a near optimal configuration of subdomains where each is placed on its own Gaussian peak, as shown in the bottom row of Figure \ref{fig:poisson2}. On the other hand, local adaptivity alone of the subdomains results in some Gaussian peaks being covered by multiple subdomains while other peaks don't have a dedicated subdomain, as shown in the middle two rows of Figure \ref{fig:poisson2}. This example showcases the benefits of residual-based subdomain addition for problems with localized features.

\subsubsection{Allen--Cahn Equation}\label{subsec:ac}
In this example, we demonstrate how the residual-based subdomain addition strategy can be used to improve the convergence of a PINN which is either exhibiting slow convergence or has gotten stuck in a local minima.  Typically, one stops training when a PINN has reached a local minimum, tunes several hyperparameters, and then attempts to train the model again from scratch. With our residual-based subdomain addition, we instead retain the progress of the PINN which has struggled to converge and we improve the solution with the help of new AB-PINN+ subdomains. In practice, this simply corresponds to training an AB-PINN+ model in which the initial PINN is regarded as the global network.

\begin{figure}[h!]
  \centering
  
  \subfloat[PDE residual loss and absolute error of the predicted solution. The curves spike every time a new subdomain is added, which we highlight with red \texttt{x} markers.]
  {\includegraphics[width=.7\textwidth]{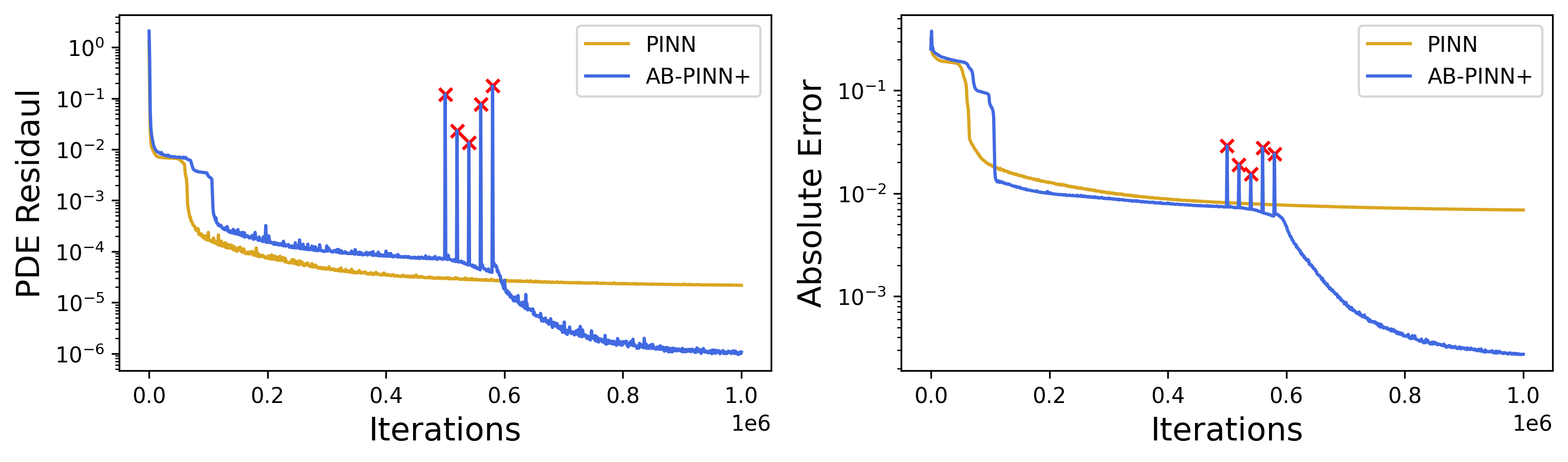}\label{fig:cahn1}}
 \\
  \subfloat[Visualization of the PINN solutions, PDE residuals, and absolute solution errors.]
  {\includegraphics[width=.85\textwidth]{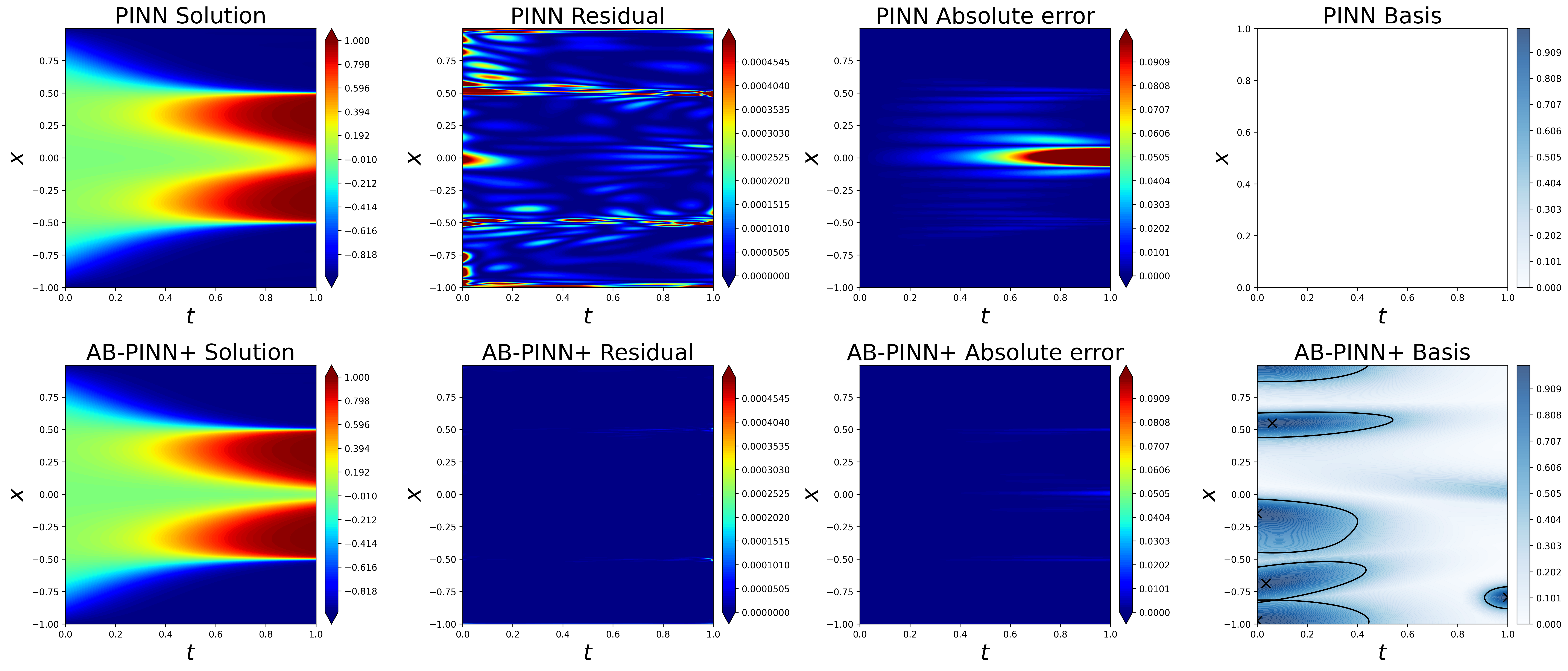}\label{fig:cahn2}} 
  \\
  \subfloat[Visualizing the slices of the predicted solution. The reference solution is plotted in black.]
  {\includegraphics[width=.9\textwidth]{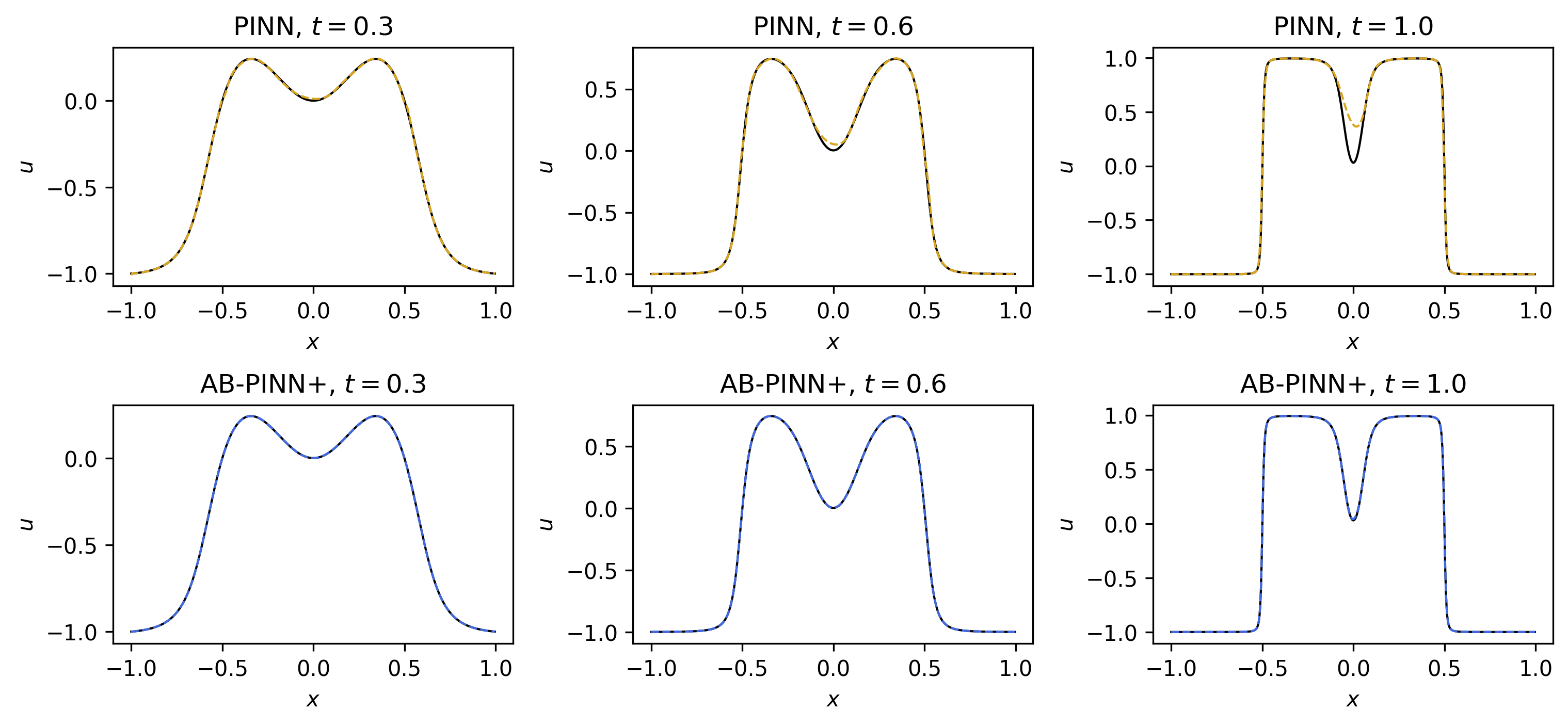}\label{fig:cahn3}}

  \caption{Using the AB-PINN+ framework to achieve a high-accuracy solution of the Allen--Cahn equation.}
  \label{fig:ac}
\end{figure}

We will consider the Allen--Cahn equation
\begin{align}
    \partial_t u(t,x) - 0.0001\cdot  \partial_{x}^2 u(t,x) + 5 u^3(t,x) - 5 u(t,x) = 0&,\qquad (t,x)\in (0,1)\times (-1,1),\label{eq:ac1} \\
    u(0,x) = x^2 \cos(\pi x)&,\qquad x\in (-1,1), \label{eq:ac2}\\
    u(t,1)=u(t,-1),\quad  u_x(t,1) = u_x(t,-1)&, \qquad t\in (0,1), \label{eq:ac3}
    \end{align}
which is a standard benchmark problem in the PINN literature. The reference solution to \eqref{eq:ac1}-\eqref{eq:ac3} is shown in Figure \ref{fig:cahnref}. We begin by solving the system \eqref{eq:ac1}-\eqref{eq:ac2} using a standard  PINN with four hidden layers and 10 nodes per layer which imposes the periodic boundary conditions \eqref{eq:ac3} using Fourier embedding (see Section \ref{subsec:periodicity}), and the initial condition \eqref{eq:ac2} with the constraining operator 
$$\Psi[u](t,x) = \tanh(t) u(t,x) + x^2 \cos(\pi x).$$

The PINN we consider has approximately 400 tunable parameters, and thus its convergence slows significantly before a high-accuracy prediction of the PDE is achieved. After $5\cdot 10^5$ iterations have elapsed, we begin adding new AB-PINN+ subdomains every $2\cdot 10^4$ iterations until $N = 5$  subdomains have been introduced, while the original PINN continues training as the global network. Each subdomain is initialized with a four hidden layer subnetwork also comprising of 10 nodes per layer, and we set $L_i = \text{diag}(1.5,0.75,0.75)$ which is defined in the three-dimensional Fourier embedded space. After all AB-PINN+ subdomains have been added, the resulting model has approximately $2.3\cdot 10^3$ tunable parameters. The learning rates for the window function parameters $\mu_i$ and $L_i$ are initialized at $10^{-3}$ and $10^{-2}$, respectively, and they are frozen after $6.5\cdot 10^5$ iterations. 

In Figure \ref{fig:cahn1}, we see that both the PDE  residual loss and absolute error of the predicted solution spike every time a new subdomain is added. This makes intuitive sense, as we significantly perturb the solution every time a new subdomain is introduced. In total, we see five spikes in Figure \ref{fig:cahn1}, each corresponding to the addition of a new subdomain. After the new window functions and corresponding subnetworks have been added, the residual loss and absolute error reduce rapidly. The final predicted AB-PINN+ solution then shows significant improvement compared to the solution without subdomain addition; see Figures \ref{fig:cahn2} and \ref{fig:cahn3}.

\subsubsection{Korteweg--De Vries Equation}\label{subsec:kdv}
We conclude our numerical experiments with a challenging PINN benchmark example, the Koreteweg--De Vries (KdV) equation:
\begin{align}
    \partial_t u(t,x) + \eta \cdot u(t,x) \partial_x u(t,x) + \mu^2 \cdot \partial_x^3 u(t,x) = 0& ,\qquad (t,x)\in (0,1)\times (-1,1) \label{eq:kdv1} \\ 
    u(0,x) = \cos(\pi x)&, \qquad x\in [0,1]\label{eq:kdv2} \\
    u(t,1) = u(t,-1)&, \qquad t\in (0,1)\label{eq:kdv3}.
\end{align}
The KdV equation is used to model nonlinear dispersive waves and can feature a range of multiscale behaviors. Here, we consider parameter values $\eta=1$ and $\mu=0.022$ leading the initial cosine to evolve into a train of solitary-type waves \cite{zabusky1965interaction,wang2024piratenets}; see Figure \ref{fig:kdvref} for a visualization of the reference solution. 

We attempt to solve  \eqref{eq:kdv1}-\eqref{eq:kdv2} using both a standard PINN and an AB-PINN+. The initial condition~\eqref{eq:kdv2} is enforced using the constraining operator 
$$\Psi[u](t,x) = \tanh(t) u(t,x) + \cos(\pi x),$$
and the periodic boundary condition \eqref{eq:kdv3} is enforced using Fourier embeddings; see Section \ref{subsec:periodicity}.

\begin{figure}[h!]
  \centering
  
  \subfloat[PDE residual loss and absolute error of the predicted solution. The red \texttt{x} indicates when new subdomains are introduced.]
  {\includegraphics[width=.7\textwidth]{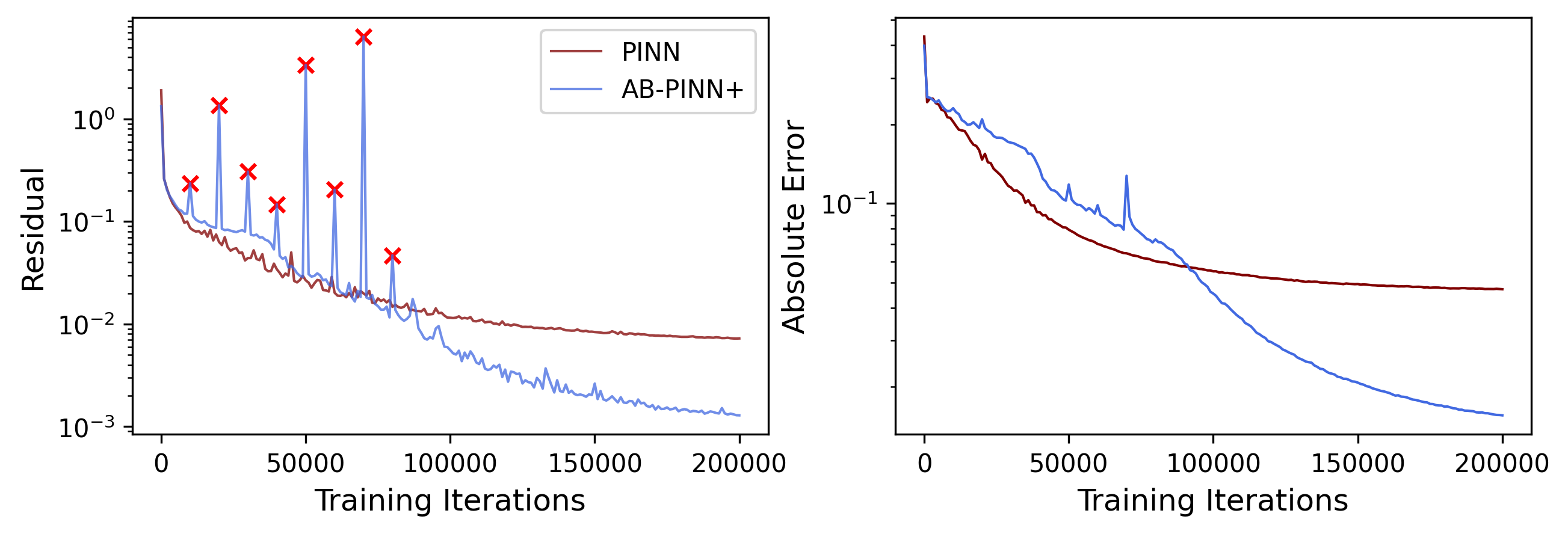}\label{fig:kdv1}}
 \\
  \subfloat[Visualization of the PINN solution, PDE residual, absolute solution error, and the learned window functions.]
  {\includegraphics[width=.9\textwidth]{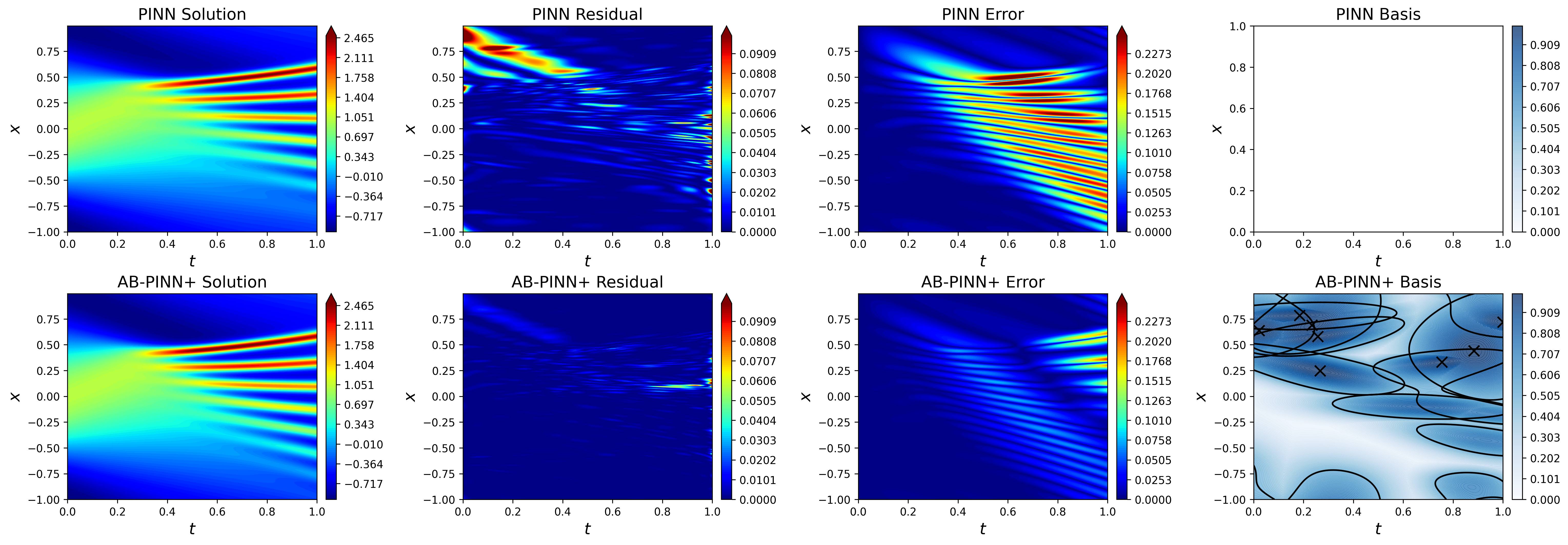} \label{fig:kdv2}}
  \\
  \subfloat[Visualizing slices of the predicted PINN solutions. The reference solution is plotted in black.]
  {\includegraphics[width=.9\textwidth]{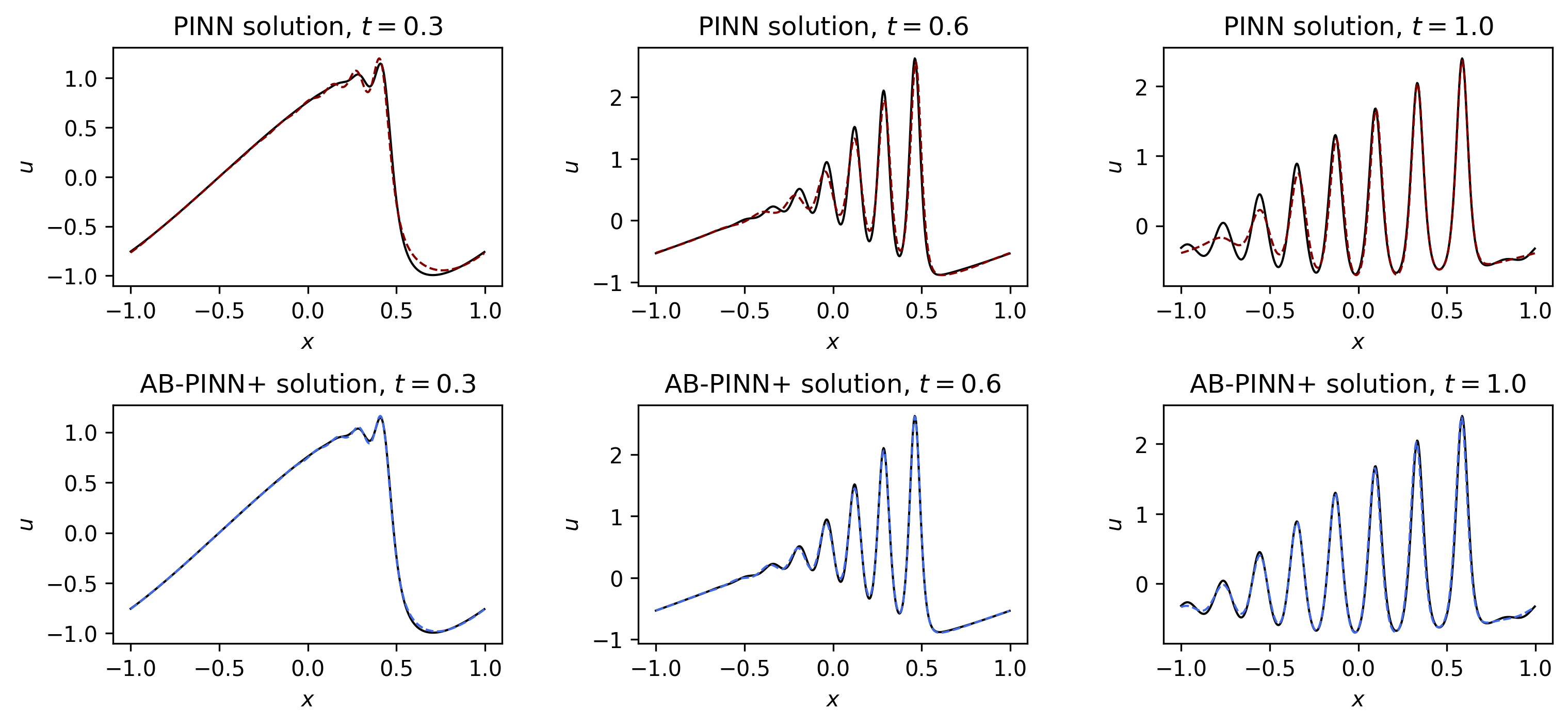} \label{fig:kdv3}}

  \caption{Solving the KdV equation using a standard PINN and an AB-PINN+.}
  \label{fig:kdv}
\end{figure}

In Section \ref{subsec:ac}, we showed how residual-based subdomain addition can prevent PINNs from getting stuck in local minima by adding additional expressive power in regions of high residual loss. Thus, the final AB-PINN+ architecture in Section \ref{subsec:ac} was larger than the standard PINN we used for comparison. Here, we instead compare the final AB-PINN+ model with a standard PINN of equal size. That is, the primary difference in experimental setup with Section \ref{subsec:ac} is simply the size of the standard PINN we compare with.

In particular, all models considered have roughly $5\cdot 10^3$ tunable parameters. The PINN uses four hidden layers with 41 nodes per layer, while the AB-PINN+ uses a four hidden layer global network with 25 nodes per layer and four hidden layer subnetworks with 10 nodes per layer. We train both a standard PINN and an AB-PINN+ for $2\cdot 10^5$ iterations, randomly sampling 1000 collocation points each step. The tunable window function parameters $\mu_i$ and $L_i$ have learning rates initialized as $10^{-3}$ and $5\cdot 10^{-3},$ respectively, and they are frozen at $10^5$ iterations. The parameters $L_i$ are initialized in the same way as in Section \ref{subsec:ac}. We use residual-based subdomain addition to introduce a new subdomain every $10^4$ iterations, until $N = 8$ subdomains are active.

As shown in Figure \ref{fig:kdv}, the AB-PINN+ model using residual-based subdomain addition approximates the reference solution of the KdV equation with higher accuracy than the standard PINN. Similar to Figure \ref{fig:cahn1}, we see that the loss spikes each time a new subdomain is added. After the subdomains have been added, the AB-PINN+ converges with lower PDE residual and absolute solution error than the standard PINN. In Figure \ref{fig:kdv2}, we observe that a large cluster of subdomains is placed in the upper-left corner of the spatio-temporal domain, which corresponds to a region where the standard PINN solution has particularly high PDE residual. Thus, the placement of these adaptive subdomains is likely responsible for the AB-PINN+'s reduction of the PDE residual in this region, contributing to a more accurate solution.

\section{Conclusion}\label{sec:conclusions}
We have introduced AB-PINNs, which provide a simple framework for dynamically learning PINN-based domain decompositions for multiscale PDEs. Building off of the FBPINN framework for static PINN-based domain decompositions \cite{moseley2020solving}, our approach provides three main new contributions. First, the parameters of each FBPINN subdomain are made learnable during training and are updated via gradient-based methods. Thus, the domain decomposition evolves throughout training and adapts to the structure of the underlying PDE solution. Second, we introduce new adaptive subdomains in regions of high PDE residual. This effectively delivers localized expressive power in regions of the spatio-temporal domain where the solution of the PDE is challenging to represent and can help prevent PINNs from getting stuck in unwanted local minima. Finally, we include a global network which is primarily responsible for propagating information between subdomains and learning the low frequency behavior of the underlying solution. We have presented numerical experiments on a wide range of PDEs which collectively show that AB-PINNs can offer significant advantages compared to both standard PINNs and  FBPINNs.

\section*{Acknowledgments}
This work was partially completed while J. Botvinick-Greenhouse interned at Mitsubishi Electric Research Laboratories (MERL). J. Botvinick-Greenhouse was also supported in part by a fellowship award under contract FA9550-21-F-0003 through the National Defense
Science and Engineering Graduate (NDSEG) Fellowship Program, sponsored by the Air Force Research Laboratory
(AFRL), the Office of Naval Research (ONR) and the Army Research Office (ARO). S. Mowlavi and W. H. Ali were supported solely by MERL. This work was done prior to M. Benosman joining Amazon Robotics.
\bibliographystyle{unsrt}  
\bibliography{references}

\end{document}